\documentclass[HARVARD,doublespace]{USG}

\usepackage{anyfontsize}

\usepackage{colortbl}   
\usepackage{xcolor}     

\graphicspath{{Figures/}{images/}}

\articletype{ORIGINAL ARTICLE}

\usepackage{macros}

\usepackage{booktabs}

\algrenewcommand{\textproc}[1]{#1}
\usepackage{setspace}

\usepackage[mathlines]{lineno}

\articledoi{10.1002/0000}

\begin{document}


\title{Accounting for model error in ensemble variational data assimilation through learned generative dynamics}

\author[1]{Marc Bocquet}[https://orcid.org/0000-0003-2675-0347]
\author[1]{Tobias S. Finn}[https://orcid.org/0000-0001-9585-8349]
\author[1]{Sibo Cheng}

\address[1]{CEREA, ENPC, EDF R\&D, Institut Polytechnique de Paris, \^Ile-de-France, France}

\corres{Marc Bocquet (\email{marc.bocquet@enpc.fr})}


\abstract[ABSTRACT]{
Many approaches have recently been proposed to combine data assimilation with machine learning for improved forecasts
with geophysical models. Here, we focus on combining advanced sequential data assimilation schemes with a generative
forecasting model which accounts for complex model errors. This generative model provides a stochastic approximation of
the conditional distribution of the successor state given the current state. The proposed variational formulation is a
novel one-step weak-constraint 4D-Var scheme in which a learned conditional stochastic forecast replaces the prescribed
additive Gaussian model error representation.  We derive and implement this formulation as an ensemble variational
algorithm, the \emph{generative iterative ensemble Kalman filter} (IEnKF-G), which generalises the iterative ensemble
Kalman filter with additive Gaussian model error (IEnKF-Q). We extend the method to a smoother with an assimilation
window spanning multiple time steps and propose a simplified variant.

The IEnKF-G is evaluated on a stochastic variant of the Lorenz 1963 model to assess its robustness. It is then tested on
a multi-scale variant of the one-dimensional Lorenz 1996 model, in which the resolved variables are affected by model
error originating from the hidden small-scale variables. The algorithm is then compared to an ensemble Kalman filter
that adaptively mitigates model and sampling errors, and to the IEnKF-Q. The IEnKF-G is generally more accurate in the
tested configurations, with substantial gains under partial observations and significant model error. This demonstrates
the potential of generative data-driven models when paired with classical data assimilation methods, particularly when
model error is substantial.
}

\maketitle



\section{Introduction}

Nowadays, machine learning enhances data assimilation (DA) in many ways to improve state estimation and forecasting
within geophysical models \citep{cheng2023}. One of the most straightforward approaches consists in using a deep
learning surrogate of the dynamics within a classical data assimilation scheme. This can accelerate forecasts and
provide tangent linear and adjoint models for variational DA
\citep[e.g.,][]{farchi2021a,farchi2025,durand2025,solvik2025}. In this context, these emulators are usually
deterministic, needing external methods to quantify their uncertainty. In contrast, generative emulators
\citep{finn2024b,price2025} offer a probabilistic representation of the forecast dynamics. Recently, they have been
incorporated into sequential DA through particle filtering \citep{savary2026}.

Here, we investigate their integration into iterative ensemble variational DA which may be scaled up to high-dimensional
systems. Our approach treats the initial state and a set of parameters driving the generative forecast model as joint
control variables, which are inferred from observations. This yields the \emph{generative iterative ensemble Kalman
filter} (IEnKF-G), which uses the learned stochastic dynamics to represent model uncertainty and which relies on
observations to constrain its realisation within the assimilation window of each DA cycle. Towards that goal, we first
briefly describe state-of-the-art \emph{sequential} DA methods that can handle model error, possibly relying on machine
learning to do so.

\subsection{Model error in data assimilation}

  
Within ensemble DA, a pragmatic way to handle model error is to compensate for neglected uncertainty by increasing the
ensemble spread. Multiplicative inflation rescales ensemble anomalies or covariances, while additive inflation injects
random perturbations, often drawn from climatological differences, stochastic physics, or model error statistics
\citep[e.g.,][]{anderson1999,anderson2007b,whitaker2012,raanes2019a}. These mitigation methods are robust, inexpensive
at inference, and widely used, but they primarily compensate for uncertainties due to model errors without explicitly
estimating their dynamical structure.

  
A more explicit approach is to augment the forecast model with an additive controllable error term, whose statistics are
described by a model error covariance matrix $\bQ$. Weak-constraint 4D-Var does this by relaxing the perfect-model
constraint and estimating, jointly with the state, a sequence of model error forcings
\citep[e.g.,][]{sasaki1970,derber1989,zupanski1997,tremolet2006,tremolet2007}. In the ensemble variational setting, the
iterative ensemble Kalman filter and smoother with additive Gaussian noise (IEnKF-Q and IEnKS-Q, respectively) extend
those methods to additive model noise by augmenting propagated analysis anomalies with model-noise anomalies in the
ensemble subspace \citep{sakov2018b,fillion2020}.

  
Instead of prescribing a stochastic additive error, hybrid correction methods seek to learn a statistical closure term
that corrects the physical model tendencies \citep[e.g.,][]{brajard2020,bocquet2020,farchi2023}. In neural network
4D-Var formulations, the correction is embedded in the forecast model and trained either offline from analysis
trajectories or online within the variational assimilation cycle, so that the DA system supplies both state estimates
and model error information \citep[e.g.,][]{farchi2025}. This strategy is attractive because it links model error
correction to model development, but it also raises identifiability issues, since analysis increments can mix model
error, observation bias, representativeness error, and DA-system deficiencies.


Score-based DA uses a neural network to represent the score, i.e. the gradient of the log-density of state or trajectory
distributions, and then conditions the reverse generative process on observations. In contrast to Gaussian-based
ensemble Kalman filters (EnKFs) or quadratic variational formulations, this offers a route to nonlinear and non-Gaussian
posterior sampling, with the observation model possibly imposed only at inference time \citep{rozet2023a}. Recent
developments include applications to geophysical trajectories and quasi-geostrophic models \citep{rozet2023b}, as well
as ensemble score filters and latent score filters for high-dimensional or sparsely observed systems
\citep{bao2024,si2025}. Since the learned trajectory prior implicitly encodes the stochastic dynamics, score-based DA
also partially replaces model error statistics with a learned distribution. In contrast to score-based DA, which
conditions the generative model by guiding its reverse sampling dynamics, the IEnKF-G retains a classical ensemble
variational scheme and exerts some control over the generative forecast model from the observations.

\subsection{Outline}

In Sect.~\ref{sec:theory}, we first recall the Bayesian principles of the iterative ensemble Kalman filters (IEnKFs),
then discuss the introduction of a probabilistic forecast model. On these grounds, we develop the IEnKF-G, which
generalises the IEnKFs to this probabilistic forecast model with the aim of estimating and counteracting model error. A
smoother derived from the IEnKF-G is also introduced.  In Section~\ref{sec:numerics}, we numerically evaluate the
IEnKF-G with low-order chaotic dynamics using a forecast model with model error. The IEnKF-G is compared against
state-of-the-art DA methods which mitigate or combat model error. Section~\ref{sec:conclusion} adds further discussion
and concludes.

\section{Theory}
\label{sec:theory}

In this theoretical section, we first formulate sequential DA in a Bayesian framework and recall the principles of
iterative ensemble Kalman filters (IEnKFs). We then introduce a generative approximation of the conditional transition
law of the assimilated state, allowing for uncertainty associated with unresolved variables or stochastic dynamics. On
this basis, we derive the IEnKF-G, describe its ensemble implementation, and discuss its longer time window counterpart.

The DA problem is formulated from the forecast and observation equations:
\begin{subequations}
  \begin{align}
    \label{eq:da-mod}
    &\bx_{k+1} = \Mc(\bx_k,\errok), \\
    \label{eq:da-obs}
    &\by_k = \Hc_k(\bx_k)+\bxi_k,
  \end{align}
\end{subequations}
respectively. In Eq.~\eqref{eq:da-mod}, $\Mc$ is an \emph{autonomous} chaotic evolution model that depends on the state
vector $\bx_k\in \Ex=\R^{\Nx}$ at time $t_k$. The vector $\errok\in \Ee=\R^{\Nerr}$ \emph{parametrises} model error at
$t_k$ and will hereafter be called the \emph{seed}. In Eq.~\eqref{eq:da-obs}, $\by_k$ is the observation vector whose
size may depend on $k$, and $\Hc_k$ is the observation operator. The observation errors $\bxi_k$ are simply assumed to
be additive, unbiased, white in time, and Gaussian with covariance matrix $\bR_k$, since observation errors are not the
focus of this work. Moreover, the sequences $\left\{\errok\right\}_k$ and $\left\{\bxi_k\right\}_k$ are assumed
independent of each other. For simplicity, we assume a constant duration $\Delta_t=t_{k+1}-t_k$ between observation
times.

\subsection{Bayesian sequential data assimilation}

Our primary objective is to estimate the conditional filtering probability density function (pdf) $p(\bx_k|\by_{:k})$,
where $\bx_k$ is the state vector at time $t_k$, and $\by_{:k}$ is the collection of observation vectors up to $t_k$.
Sequential DA seeks a recursion in time for this conditional pdf. We first write the marginalisation identity
\begin{equation}
  \label{eq:cond-pdf}
  p(\bx_k|\by_{:k}) = \int_{\Ex}\! {\mathrm d}\bx_{k-1} \, p(\bx_k, \bx_{k-1}|\by_{:k}).
\end{equation}
Assuming Markovian state dynamics and conditionally independent observations given the states, Bayes' rule yields the recursion
\begin{equation}
  p(\bx_k|\by_{:k}) = \frac{1}{p(\by_k|\by_{:k-1})} \int_{\Ex}\! {\mathrm d}\bx_{k-1} \,
  p(\by_k|\bx_k) p(\bx_k| \bx_{k-1})p(\bx_{k-1} |\by_{:k-1}).
\end{equation}
This recursion is exact for the assumed probabilistic model.  When the resolved reference dynamics are not Markovian,
using a transition density conditioned only on the current state constitutes a Markov approximation.
Later on, it will be convenient to work on the integrand of Eq.~\eqref{eq:cond-pdf}, and to focus on the one-step
conditional pdf:
\begin{equation}
  \label{eq:one-step-filtering-k}
  p(\bx_k, \bx_{k-1}|\by_k) = \frac{1}{p(\by_k)} p(\by_k|\bx_k)p(\bx_k|\bx_{k-1})p(\bx_{k-1}),
\end{equation}
with an implicit dependence on $\by_{:k-1}$ for the sake of brevity. Focusing hereafter on the generic interval $[t_0,
  t_1]$ to avoid handling the $k$ subscripts, Eq.~\eqref{eq:one-step-filtering-k} equivalently reads
\begin{equation}
    \label{eq:one-step-filtering}
  p(\bx_1, \bx_0|\by_1) = \frac{1}{p(\by_1)} p(\by_1|\bx_1)p(\bx_1|\bx_0)p(\bx_0).
\end{equation}

\subsection{Iterative ensemble Kalman filters}

The factorisation~\eqref{eq:one-step-filtering} for the conditional pdf $p(\bx_1, \bx_0|\by_1)$ is suitable for sequential
\emph{filtering}. Yet, it can alternatively be written
\begin{equation}
    \label{eq:one-step-wc-smoothing}
  p(\bx_1, \bx_0|\by_1) = p(\bx_1|\by_1,\bx_0)p(\bx_0|\by_1),
\end{equation}
so as to emphasise the \emph{smoothing} pdf $p(\bx_0|\by_1)$. This is leveraged in the following.

\subsubsection{Perfect model assumption}

First, let us assume that the forecast model is perfectly known and deterministic so that the transition pdf from $t_0$
to $t_1$ reads $p(\bx_1|\bx_0) = \delta\lp\bx_1-\Mc(\bx_0)\rp$. Hence, appearing as the first pdf on the right-hand side
of Eq.~\eqref{eq:one-step-wc-smoothing} we have
\begin{equation}
  \label{eq:det-transition}
  p(\bx_1|\by_1,\bx_0)=\delta\lp\bx_1-\Mc(\bx_0)\rp,
\end{equation}
and, for the second pdf on the right-hand side of Eq.~\eqref{eq:one-step-wc-smoothing}, we have
\begin{equation}
  \label{eq:smoothing-pdf}
  p(\bx_0|\by_1) = \frac{1}{p(\by_1)}p(\by_1|\bx_0)p(\bx_0) = \frac{1}{p(\by_1)} p(\by_1|\Mc(\bx_0)) p(\bx_0).
\end{equation}
Equation~\eqref{eq:smoothing-pdf} suggests first solving the smoothing problem (before the filtering one) based on the
cost function
\begin{equation}
  \label{eq:ienkf-cf-1}
  -\ln p(\bx_0|\by_1) = -\ln p(\by_1|\Mc(\bx_0)) - \ln p(\bx_0)  + \mathrm{cst},
\end{equation}
so as to estimate a mode $\bx_0^\star$ of the smoothing posterior.  In the ensemble implementation, a smoothing ensemble
is constructed around this estimate and propagated through $\Mc$; the filtering estimate is then given by the mean of
the propagated ensemble.  This is the Bayesian rationale of the iterative ensemble Kalman filter
\citep[IEnKF,][]{sakov2012}, whose detailed Bayesian derivation can be found in \citet{bocquet2014a}. Under the
idealised perfect model assumption, and applied to chaotic dynamics, it improves both smoothing and filtering accuracy
as compared to the ensemble Kalman filter or smoother \citep{bocquet2013}.

Assuming the prior and observation error distributions are Gaussian, i.e.
\begin{subequations}
    \label{eq:gauss-err-1}
  \begin{align}
  -\ln p(\by_1|\bx_1) &= \frac{1}{2}\left\| \by_1 - \Hc(\bx_1) \right\|_{\bRi}^2 + \mathrm{cst}, \\
  -\ln p(\bx_0) &= \frac{1}{2}\left\| \bx_0 - \xa_0\right\|_{\lp\Pa_0\rp^{-1}}^2 + \mathrm{cst},
\end{align}
\end{subequations}
where $\left\| \bx \right\|^2_{\bP} = \bx\T\bP\bx$, $\bR$ is the (positive definite) observation error covariance matrix
at $t_1$, and $\Pa_0$ is the (positive definite) background error covariance matrix at $t_0$.
The IEnKF cost function associated with Eq.~\eqref{eq:ienkf-cf-1} then becomes
\begin{equation}
  \label{eq:ienkf-cf-2}
  -\ln p(\bx_0|\by_1) = \frac{1}{2}\left\| \bx_0 -
  \xa_0\right\|_{\lp\Pa_0\rp^{-1}}^2 + \frac{1}{2}\left\| \by_1 - \Hc\circ\Mc(\bx_0) \right\|_{\bRi}^2  +  \mathrm{cst},
  \end{equation}
which coincides with a one-step cost function of strong-constraint 4D-Var. The vanilla IEnKF's analysis then relies on
an ensemble implementation of a Gauss--Newton minimisation of Eq.~\eqref{eq:ienkf-cf-2}.

\subsubsection{Imperfect model with Gaussian noise}

Second, assume instead that the model is imperfect with Gaussian model noise. Accordingly, the cost associated
with the transition pdf $p(\bx_1|\bx_0)$ reads
\begin{equation}
  \label{eq:gauss-err-2}
  -\ln p(\bx_1|\bx_0) = \frac{1}{2}\left\| \bx_1-\Mc(\bx_0)\right\|^2_{\bQ^{-1}} + \mathrm{cst},
\end{equation}
with $\bQ$ the (positive definite) Gaussian model error covariance matrix.
In that case, the associated cost function can be obtained from Eq.~\eqref{eq:one-step-filtering} and the error
statistics assumptions Eqs.~\eqref{eq:gauss-err-1} and~\eqref{eq:gauss-err-2}:
\begin{equation}
  \label{eq:gauss-err-3}
  -\ln p(\bx_1,\bx_0|\by_1) = \frac{1}{2}\left\| \bx_0-\xa_0\right\|^2_{\lp\Pa_0\rp^{-1}}
 + \frac{1}{2}\left\| \by_1-\Hc\lp\bx_1\rp\right\|^2_{\bRi} 
+   \frac{1}{2}\left\| \bx_1-\Mc(\bx_0)\right\|^2_{\bQ^{-1}} + \mathrm{cst},
  \end{equation}
which coincides with the cost function of a one-step weak-constraint 4D-Var. This cost function is the backbone of the
IEnKF-Q \citep{sakov2018b}. Alternatively, especially when the observation operator is linear, it can be more efficient
to rely on the alternative expression Eq.~\eqref{eq:one-step-wc-smoothing}. Detailed derivations, including
ensemble-based Gauss--Newton implementations of Eq.~\eqref{eq:gauss-err-3}, are given in
\citet{sakov2018b,sakov2018a,fillion2020}.

\subsection{The generative iterative ensemble Kalman filter}

\subsubsection{Imperfect model}

Now, let us consider more general transition pdfs for $p(\bx_1|\bx_0)$ and discuss the core of the
main proposal of this paper. Let us assume that the transition pdf can be written as
\begin{equation}
  \label{eq:ienkfg-ck-pdf}
   p(\bx_1|\bx_0) = \int\!\mathrm{d}\err \, \delta\lp\bx_1-\Mc(\bx_0,\err)\rp p_\err(\err),
\end{equation}
where $\err \mapsto p_\err(\err)$ is the standard Gaussian pdf, and $\bx_0,\err \mapsto \Mc(\bx_0,\err)$ is the
generative forecast map over $[t_0,t_1]$.  In particular, for a fixed initial state, varying the seed $\err$ generates
different possible successor states. In the subsequent section, we discuss how to define and obtain such a generative
map.

\subsubsection{A probabilistic surrogate for the transition probability density}

The transition pdf, $p(\bx_1|\bx_0)$, or more precisely a numerically efficient procedure to sample from it, can be
obtained with a generative machine learning model trained on successive-state pairs $\left\{ (\bx_k,\bx_{k+1})
\right\}_{k=1,\ldots,K}$ of the dynamics. The samples in such a dataset can be collected from a long trajectory assuming
ergodicity or from a batch of shorter trajectories.

Suitable generative methods include denoising diffusion \citep{ho2020,karras2022}, flow-matching \citep{lipman2023}, or
stochastic interpolants \citep{chen2024b}, in order to learn the stochastic dynamics that allow us to sample from the
transition pdf $p(\bx_1|\bx_0)$. Irrespective of the details of their implementation, these methods can be seen as
yielding a map $\Mc$,
\begin{equation}
  \Mc: (\bx_0 ,\err)\in\Ex\oplus\Ee \mapsto \bx_1=\Mc(\bx_0 ,\err)\in\Ex,
\end{equation}
which is related to $p(\bx_1|\bx_0)$ through Eq.~\eqref{eq:ienkfg-ck-pdf}.  Note that this definition of $\Mc$ also
matches the training of stochastic surrogate dynamics using loss functions based on ensemble scores such as the
continuous ranked probability score (CRPS) or the energy score
\citep{gneiting2007,gneiting2008,pacchiardi2024,lang2026}. Such a map has also been called \emph{engression} by
\citet{shen2025}.  Marginalised over the seed $\err$, $\Mc$ can be understood as stochastic dynamics, the many images of
$\bx_0$ being parametrised by $\err$.  We include the seed $\err$ among the DA control variables and optimise it jointly
with the initial state. This requires suitable smoothness of $\Mc$ with respect to both arguments.

The particle filter approach of \citet{savary2026} also exploits a learned transition distribution, but performs
inference through observation-conditioned proposals and particle weighting. Here, inference is formulated as an
optimisation problem in the augmented state--seed ensemble space.

The seed $\err$ parametrises variability in the learned conditional distribution of the successor state given the
current state. This variability can have different origins. If the tracked state omits dynamically relevant variables,
its successor need not be uniquely determined by its current value, even when the full dynamics are
deterministic. Intrinsically stochastic dynamics also generate conditional variability, including when their transition
law is known exactly. For fully resolved deterministic Markov dynamics, however, the true conditional transition law is
a Dirac measure. Moreover, finite training data, limited model capacity and imperfect optimisation can lead to a learned
transition law that differs from this target, possibly including a non-zero spread. Such spread should not automatically
be interpreted as calibrated uncertainty about the unknown dynamics \citep{chan2024}. More generally, variability
generated by a fixed trained model does not by itself quantify uncertainty in the learned transition law.

The seed $\err$ can be defined in the ad hoc parameter space $\Ee=\R^{\Nerr}$. Note that $\err$ is often defined in
state space, i.e. $\Ee=\Ex$. Nonetheless, $\err$ can be made to span a subspace of $\Ex$, and would hence be represented
by fewer degrees of freedom than $\Nx$.

For what follows, the most important assumption is that, in those schemes, the training assumes that $\err$ is
independently sampled from the standard Gaussian distribution: $\err \sim \gauss (\bzero,\bI_{\Nerr})$, with the
corresponding pdf
\begin{equation}
  p_{\err}(\err)=n(\err; \bzero,\bI_{\Nerr}),
\end{equation}
where $n$ is the Gaussian pdf and $\bI_m$ hereafter stands for the identity matrix in $\R^m$. This Gaussian distribution
is a natural prior for $\err$ and will be leveraged in the description of the IEnKF-G.

\subsubsection{Bayesian rationale and cost function}

Since $p(\bx_1|\bx_0)$ is represented by Eq.~\eqref{eq:ienkfg-ck-pdf}, we could use Eq.~\eqref{eq:one-step-filtering}
and its negative logarithm to obtain a cost function.  However, Eq.~\eqref{eq:ienkfg-ck-pdf} involves an integral which
is inconvenient to evaluate.  That is why we will describe the DA problem in terms of the variables $(\bx_0,\err)$
rather than $(\bx_0,\bx_1)$ and work with the associated one-step conditional pdf, rather than
Eq.~\eqref{eq:one-step-filtering}:
\begin{subequations}
  \label{eq:ienkfg-pdf}
  \begin{align}
    \label{eq:ienkfg-pdf-a}
    p(\bx_0, \err|\by_1) &= \frac{1}{p(\by_1)} p(\by_1|\bx_0,\err) p(\bx_0,\err) \\ 
    &=  \frac{1}{p(\by_1)} p(\by_1|\Mc(\bx_0,\err)) p(\bx_0) p_\err(\err),
  \end{align}
\end{subequations}
where $\err$ is assumed a priori independent from $\bx_0$ and past observations.
Equation~\eqref{eq:ienkfg-pdf} offers the Bayesian rationale for the IEnKF-G.
With the aforementioned Gaussian assumptions on the errors and the seed $\err$, its associated cost function is defined by
$ -\ln p(\bx_0,\err|\by_1)$ up to terms independent of $\bx_0$ and $\err$ and reads
\begin{equation}
  \label{eq:ienkfg-cf}
  \Lc(\bx_0,\err) = \frac{1}{2}\left\| \bx_0-\xa_0\right\|^2_{\lp\Pa_0\rp^{-1}}
 + \frac{1}{2}\left\| \by_1-\Hc\circ\Mc(\bx_0,\err)\right\|^2_{\bRi}  + \frac{1}{2}\left\| \err\right\|^2.
\end{equation}

In the case where $\Nerr=\Nx$, the model error is Gaussian, i.e. $\Mc(\bx_0,\err)=\Mc(\bx_0) + \bQ^{1/2}\err$, and
$\bQ$, the (positive definite) model error covariance matrix has no dependence on $\bx_0$,
the latent representation $(\bx_0,\err)$ is not needed to evaluate this transition density
since Eq.~\eqref{eq:ienkfg-ck-pdf} can be straightforwardly integrated out:
\begin{equation}
  p(\bx_1 | \bx_0) = n\lp\bx_1; \Mc(\bx_0), \bQ\rp.
\end{equation}
This leads back to the Bayesian foundation of the IEnKF-Q; hence, the IEnKF-G is a generalisation of the IEnKF-Q.

\subsubsection{Ensemble subspace}

Following the derivations of, e.g., \citet{sakov2018b} for the IEnKF-Q, we now develop an ensemble method for the
IEnKF-G from the cost function Eq.~\eqref{eq:ienkfg-cf}. We use an ensemble-based reduction strategy of the error
covariances similar to that of the IEnKF and IEnKF-Q, with an ensemble size $\Ne$. As usual in ensemble methods, the
empirical Gaussian prior is understood as a measure supported on the affine ensemble subspace and represented through
standard ensemble coordinates (denoted $\bu$ hereafter), so that the possibly singular state-space covariance need not
be inverted \citep{hunt2007}. We assume the ensemble parametrisation:
\begin{equation}
  \bx_0 = \xa_0 + \Xa_0\ue,
\end{equation}
for the state vectors where $\ue\in\R^{\Ne}$.  The matrix $\Xa_0\in\Ex^{\Ne}$ is assumed to satisfy
$\Xa_0\bone_{\Ne}=\bzero$ and $\Pa_0=\Xa_0\lp\Xa_0\rp\T$, which is traditional for ensemble transform-based filters.
Hence, $\Xa_0$ is a matrix of normalised centred anomalies.  For the seed, we assume
\begin{equation}
  \err = \Xg_0\ug,
\end{equation}
where $\ug\in\R^{\Ng}$ and $\Xg_0\in\Ee^{\Ng}$ is a centred matrix of seed anomalies, $\Xg_0\bone_{\Ng}=\bzero$, whose
normalisation is specified below. Hence, the seed has up to $\min(\Ng-1,\Nerr)$ degrees of freedom at each time step.

\subsubsection{Seed subsampling, cycling, and statistical consistency}

For $\Ng \ge 2$, assume the rank of $\Xg_0$ is $m=\min(\Ng-1,\Nerr)$. We normalise $\Xg_0$ such that
\begin{equation}
  \label{eq:seed-normalisation}
  \Xg_0(\Xg_0)\T=\frac{\Nerr}{m}\bPi_m,
\end{equation}
where $\bPi_m$ is the orthogonal projector onto the column space of $\Xg_0$. The orientation of this $m$-dimensional
subspace is drawn isotropically at random and will be redrawn independently at every assimilation cycle (relying on
Algorithm~\ref{alg:regular-simplex} in Appendix~\ref{app:samplers}).  Equation~\eqref{eq:seed-normalisation} fixes the
total seed variance to $\operatorname{Tr}[\Xg_0(\Xg_0)\T]=\Nerr$ in every cycle. Since the orientation is isotropic,
$\mathbb E[\Xg_0(\Xg_0)\T]$ is proportional to $\bI_{\Nerr}$, and the trace constraint then yields $\mathbb
E[\Xg_0(\Xg_0)\T]=\bI_{\Nerr}$: the construction is an unbiased, randomised low-rank representation of the seed prior
covariance. Within one cycle, the seed is confined to an $m$-dimensional subspace with variance $\Nerr/m$ per direction.
Over the cycles, the successive subspaces sweep the whole seed space, and the average of $\Xg_0(\Xg_0)\T$ builds up to
$\bI_{\Nerr}$.

By contrast, restricting the seed prior to unit variance within the subspace, $\Xg_0(\Xg_0)\T=\bPi_m$, would
underestimate this contribution by a factor $m/\Nerr$ in every cycle. Let us remark that, for $2\le\Ng\le\Nerr+1$, the seed
members $\sqrt{\Ng-1}\,\Xg_0$ have squared norm $\Nerr(\Ng-1)/\Ng$ (Algorithm~\ref{alg:regular-simplex}), of the order
of the typical value $\Nerr$ for seeds drawn from $\gauss(\bzero,\bI_{\Nerr})$. This scaling keeps the initial
seed-varying perturbations at a radius comparable to the typical training seeds.

As a consequence of the statistical assumption Eq.~\eqref{eq:seed-normalisation}, the seed penalty term in the cost
function Eq.~\eqref{eq:ienkfg-cf} is modified to
\begin{equation}
  \label{eq:Xg_scaling}
   \frac{1}{2}\left\| \err\right\|^2 \longrightarrow \frac{1}{2} \err\T \left[ \Xg_0\lp\Xg_0\rp\T \right]^\dagger \err,
\end{equation}
where $\dagger$ denotes the pseudo-inverse operator.

For $\Ng=1$, the cost function reduces to that of the IEnKF driven by $\bx \mapsto \Mc(\bx, \bzero)$. Yet, the augmented
transform implementations differ and may yield a different finite-ensemble mean. Also note that $\err=\bzero$ is itself
atypical under the training distribution (see Sect.~\ref{sec:training}).

\subsubsection{Implementation}

Denoting $N=\Ne+\Ng$ and $\bu= [ \ue\T \quad \ug\T ]\T$, we obtain the parametrised state at $t_1$ as
\begin{equation}
  \bx_1(\bu) = \Mc(\xa_0+\Xa_0\ue, \Xg_0\ug),
\end{equation}
and the cost function Eq.~\eqref{eq:ienkfg-cf} with the modification Eq.~\eqref{eq:Xg_scaling} now compactly reads
\begin{subequations}
\begin{align}
  \Lc(\bu) &= \frac{1}{2}\left\| \bu \right\|^2
  + \frac{1}{2}\left\| \by_1-\Hc\lp\bx_1\rp\right\|^2_{\bRi}, \\
    \label{eq:x1_ienkfg}
 \bx_1 &= \Mc(\xa_0+\Xa_0\bu_{1:\Ne},\Xg_0\bu_{\Ne+1:\Ne+\Ng}).
\end{align}
\end{subequations}
Define the tangent linear operators (Jacobians), $\Mx = \nabla_{\bx}\Mc$, $\Me = \nabla_{\err}\Mc$, and
$\bH=\nabla_{\bx}\Hc$.  The tangent linear of $\bu \mapsto \Hc(\bx_1(\bu))$ is given by $\bH\bM\bX$ with
\begin{equation}
  \bM = \begin{bmatrix} \Mx & \Me \end{bmatrix} \in \Ex^{\Nx+\Nerr}, \qquad
  \bX = \begin{bmatrix} \Xa_0 & \bzero \\ \bzero & \Xg_0 \end{bmatrix} \in \Exe^N,
\end{equation}
where $\Exe=\Ex\oplus\Ee$ (augmented space). The smoothing state analysis $\xs_0$ is parametrised by $\bu^\mathrm{s}=
\argmin_{\bu} \Lc(\bu)$. The analysis error covariance matrix for $\bu$ can be estimated via the inverse Hessian of
$\Lc(\bu)$ approximated by first-order derivatives only:
\begin{align}
  \bD = \left[ \bI_N + (\bH\bM\bX)\T\bRi\bH\bM\bX \right]^{-1} \in \R^{N\times N}.
\end{align}
Since
\begin{equation}
  \begin{bmatrix} \bx_0-\xa_0 \\ \err \end{bmatrix} = \bX \bu,
\end{equation}
we obtain the updated smoothing normalised anomalies at $t_0$:
\begin{equation}
  \label{eq:As0_ienkfg}
  \Xs_0 = \bX \bD^{\frac{1}{2}} \in \Exe^N.
\end{equation}
$\Xs_0 \in\Exe^N$ corresponds to $N$ updated anomalies in the space $\Exe$. Its update Eq.~\eqref{eq:As0_ienkfg} is used
for smoothing but, critically, in the Gauss--Newton loop of the IEnKF-G's algorithm as well.  Note that each member of
the ensemble has its own perturbed initial state and paired model error seed.

Further, using Eq.~\eqref{eq:x1_ienkfg}, we obtain the filtering analysis normalised anomalies at $t_1$:
\begin{equation}
  \label{eq:Aa1}
  \Xa_1 = \bM \Xs_0.
\end{equation}
This linearised forecast Eq.~\eqref{eq:Aa1} actually parallels the probabilistic result:
\begin{subequations}
  \label{eq:pushforward}
\begin{align}
  p(\bx_1|\by_1) &= \int\!\mathrm{d} \bx_0 \mathrm{d} \err \, p(\bx_1,\bx_0,\err|\by_1) \\
  &= \int\!\mathrm{d} \bx_0 \mathrm{d} \err \, p(\bx_1|\bx_0,\err,\by_1) p(\bx_0,\err|\by_1) \\
  &= \int\!\mathrm{d} \bx_0 \mathrm{d} \err \, \delta\lp\bx_1-\Mc(\bx_0,\err)\rp p(\bx_0,\err|\by_1).
\end{align}
\end{subequations}
Equation~\eqref{eq:pushforward} gives the exact filtering distribution as the pushforward of the augmented posterior. In
the ensemble implementation, this posterior is approximated locally in the control coordinates and propagated to obtain
a filtering estimate and its uncertainty; the resulting estimate need not coincide with the mode of the exact filtering
distribution. $\Xa_1$ encapsulates $N=\Ne+\Ng$ updated anomalies. Since the next cycle requires $\Ne$ anomalies, $\Xa_1$
is downsampled from $N$ to $\Ne$ anomalies (see Algorithm~\ref{alg:redux-augment-rr}).  Alternatively we could compute
the marginal $p(\err|\by_1)$ so as to get an estimate of the anomalies $\Xg_1$ which would specify a new subspace for
$\err$, which could then be used to define the seed subspace of the subsequent analysis.

The main algorithm for the IEnKF-G implementing the aforementioned equations is given in Appendix~\ref{app:algorithms},
Algorithm~\ref{alg:ienkfg-transform} for the transform variant, and Algorithm~\ref{alg:ienkfg-bundle} for the bundle
one. The cycling of the IEnKF-G is illustrated by the schematic in the left panel in Fig.~\ref{fig:cycling}. For the
sake of comparison, the algorithm of the IEnKF-Q is also given in Appendix~\ref{app:algorithms},
Algorithm~\ref{alg:ienkfq} \citep[adapted from][]{sakov2018b}.

\begin{figure*}[t]
  \begin{tabular}{cc}
    \includegraphics[height=0.25\textheight]{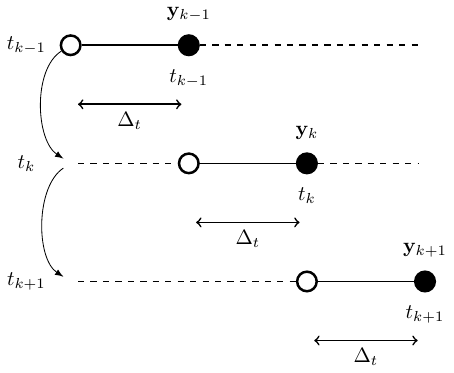} &
    \includegraphics[height=0.25\textheight]{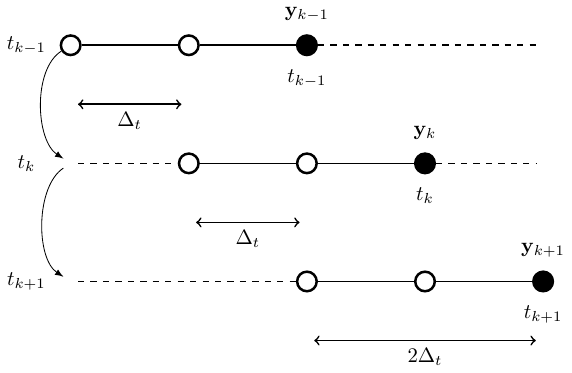}
  \end{tabular}
  \caption{ \label{fig:cycling}
    Schematic of the IEnKF-G/IEnKS-G cycling, characterised by a data assimilation window of $L$ time steps
    $\Delta_t$, and shifting the window by $S$ time steps $\Delta_t$ between analyses. The left panel corresponds to the
    IEnKF-G with $L=1$, $S=1$; the right panel corresponds to the IEnKS-G with $L=2$, $S=1$.}
\end{figure*}

The ensembles used to estimate sensitivities do not need to have the spread prescribed by the approximate posterior
covariance. For example, setting $\bSig=\bI_N$ in the transform variant as defined in
Algorithm~\ref{alg:ienkfg-transform} gives the sensitivity anomalies $\bX\bD^\frac{1}{2}$, whereas an $N$-member
ensemble representing covariance $\bX\bD\bX\T$ requires anomalies $\sqrt{N-1}\,\bX\bD^\frac{1}{2}$.
Algorithm~\ref{alg:ienkfg-transform} retains the mean of the last propagated sensitivities and reconstructs the analysis
anomalies. Algorithm~\ref{alg:ienkfg-bundle} instead performs a final propagation with the posterior spread. For
nonlinear dynamics, these constructions may yield different filtering means.

As in the IEnKF and the IEnKF-Q, the ensemble circumvents the use of the model adjoint, through either the bundle
(finite differences) or the ensemble transform (secant) approaches \citep{sakov2012,bocquet2012b}, both of which allow
us to estimate the sensitivity matrix $\bH\bM\bX$.  If the adjoint of the transition model and that of the observation
operator are available, they can be used to compute the cost function gradients rather than computing the sensitivities
with an ensemble. The ensemble sensitivity computation within the optimisation loop in the bundle variant,
Algorithm~\ref{alg:ienkfg-bundle}, can then be replaced by the explicit computation of the gradient.  In the abence of a
sensitivity ensemble, the computation of the updated ensemble would then require estimating the Hessian or its action
through tangent linear and adjoint products, or any ad hoc sampling method \citep[e.g.,][]{solvik2025}.

The computational cost depends on the combined ensemble size $N$: dense matrix factorisations cost $\mathcal{O}(N^3)$,
while each iteration requires a number of surrogate evaluations proportional to $N$. Increasing $\Ng$ therefore trades
computational cost against a richer representation of the seed variability.

Finally, note that the method also relies on a local Gaussian approximation of the augmented smoothing posterior, which
cannot represent distinct posterior modes, although nonlinear ensemble propagation can produce non-Gaussian forecast
distributions.

\subsection{The simplified generative iterative ensemble Kalman smoother}

We now consider $L\times\Delta_t$-long data assimilation windows (DAWs) with $L \ge 2$ observation batches, in contrast
to the $\Delta_t$-long DAW considered so far. The goal is to extend the IEnKF-G to longer DAWs, which would yield the
generative iterative ensemble Kalman smoother (IEnKS-G), following how the IEnKF was extended to the iterative ensemble
Kalman smoother \citep[IEnKS,][]{bocquet2014a}.  Similarly to the weak-constraint 4D-Var, the associated cost function
reads:
\begin{subequations}
\begin{equation}
  \Lc(\bx_{0:L},\err_{0:L-1}) = \frac{1}{2}\left\| \bx_0-\xa_0\right\|^2_{\lp\Pa_0\rp^{-1}}
 + \frac{1}{2}\sum_{l=1}^{L}\left\| \by_l-\Hc_l(\bx_l)\right\|^2_{\bRi_l}  + \frac{1}{2}\sum_{l=0}^{L-1}\left\|
 \err_l\right\|^2,
\end{equation}
under the constraints, for $l=0, \ldots, L-1$:
\begin{equation}
  \bx_{l+1} = \Mc(\bx_l,\err_l).
\end{equation}
\end{subequations}
The number of seed control variables grows linearly with $L$, which may become numerically prohibitive. To reduce this
cost, we use a single seed $\err$ for all transitions within the DAW \citep[e.g., Appendix~B in][]{fillion2020}, with
prior $\err\sim\gauss (\bzero,\bI_{\Nerr})$.  This replaces independent seeds by perfectly correlated seeds within the
window and therefore changes the assumed joint law of the trajectory.  The resulting optimisation problem is:
\begin{subequations}
\begin{equation}
  \Lc(\bx_{0:L},\err) = \frac{1}{2}\left\| \bx_0-\xa_0\right\|^2_{\lp\Pa_0\rp^{-1}}
 + \frac{1}{2}\sum_{l=1}^{L}\left\| \by_l-\Hc_l(\bx_l)\right\|^2_{\bRi_l}  + \frac{1}{2}\left\|\err\right\|^2,
\end{equation}
under the constraints, for $l=0, \ldots, L-1$:
\begin{equation}
  \bx_{l+1} = \Mc(\bx_l,\err).
\end{equation}
The shared seed has unit prior variance in each component, giving the single penalty $\|\err\|^2/2$. By contrast,
substituting $\err_l=\err$ into the sum of independent-seed penalties would give $L\|\err\|^2/2$, corresponding to prior
covariance $L^{-1}\bI_{\Nerr}$ for the shared seed. Lifting the constraints, the cost function can be written in the
compact constraint-free form:
\begin{equation}
  \label{eq:ienksg-cf}
  \Lc(\bx_0,\err) = \frac{1}{2}\left\| \bx_0-\xa_0\right\|^2_{\lp\Pa_0\rp^{-1}}
 + \frac{1}{2}\sum_{l=1}^{L}\left\| \by_l-\Hc_l\circ\Mc_{0:l}(\bx_0, \err)\right\|^2_{\bRi_l}  + \frac{1}{2}\left\|\err\right\|^2,
\end{equation}
\end{subequations}
where $\Mc_{0:l}$ denotes $l$ successive applications of $\Mc$ with the same seed.  When the DAW is shifted by $S$
observation intervals $\Delta_t$ ($S \le L$), successive windows overlap and, if all the observation vectors within the
window are assimilated, each one of them is assimilated $L/S$ times (assuming $L$ is a multiple of $S$). To avoid
over-counting, the observation error covariance matrices in Eq.~\eqref{eq:ienksg-cf} are then inflated,
$\bR_l\to\beta_l^{-1}\bR_l$, with weights $0<\beta_l\le1$ chosen such that, for each observation vector, the weights
applied over its successive assimilations sum to one, e.g., $\beta_l=S/L$. Following \citet{bocquet2014a}, the resulting
algorithm is called the (simplified) multiple data assimilation IEnKS-G (MDA IEnKS-G). At the end of a cycle, the most
recent observation batches have only been partially assimilated. Filtering estimates, e.g., of $\bx_L$, therefore
require an additional, non-cycled balancing analysis in which these batches are assimilated with their complementary
weights.

When $S=1$ and only the latest observation vector $\by_L$ is assimilated, the cost function becomes
\begin{equation}
  \Lc(\bx_0,\err) = \frac{1}{2}\left\| \bx_0-\xa_0\right\|^2_{\lp\Pa_0\rp^{-1}}
 + \frac{1}{2}\left\| \by_L-\Hc_L\circ\Mc_{0:L}(\bx_0, \err)\right\|^2_{\bRi_L}  + \frac{1}{2}\left\|\err\right\|^2.
\end{equation}
In this case, the DAWs virtually overlap, but the observation batches are assimilated once and only once. The algorithm
is called the (simplified) single data assimilation IEnKS-G (SDA IEnKS-G) following \citet{bocquet2014a}. The cycling of
the IEnKS-G is illustrated by the schematic in the right panel in Fig.~\ref{fig:cycling}. Seed persistence is assumed
throughout the DAW but changes from one DA cycle to the next. The algorithm pseudocode of the simplified transform SDA
IEnKS-G, i.e. in the case $S=1 \le L$, can be found in Appendix~\ref{app:algorithms},
Algorithm~\ref{alg:ienksg-transform}. For simplicity, only this variant of the IEnKS-G will be numerically evaluated.

\section{Numerical experiments}
\label{sec:numerics}

We evaluate the methods in the framework of DA twin experiments using two low-order chaotic models from which the true
trajectories to compare to are generated. The first model is based on a stochastic variant of the three-variable Lorenz
chaotic model \citep[L63;][]{lorenz1963}.  It provides an inexpensive test of IEnKF-G implementations but in challenging
nonlinear and sparse conditions, although it may not be representative of realistic configurations. It is directly
inspired by \citet{chekroun2011,savary2026}.

The second model is the Grooms--Lee 2015 model \citep[GL15;][]{grooms2015}. It can be seen as an idealised
one-dimensional high-resolution model whose fine-scale and large-scale dynamics are both based on the Lorenz 1996 (L96)
chaotic model \citep{lorenz1998}.  Unlike the two-scale Lorenz 1996 model (L96-2s) introduced by Lorenz in his ECMWF
predictability seminar paper, “Predictability: A Problem Partly Solved” \citep{lorenz1996}, which relies on an explicit
time-scale ratio (parameter $c$), GL15 has no time-scale separation parameter; its scales are separated in space only.
The reference trajectories used in the DA experiments consist of the large-scale variables of GL15. Although the full
model is Markovian, these resolved variables need not be: the unresolved variables can introduce dependence on the
past. We approximate their evolution by a transition model conditioned only on the current resolved state.  However,
learning this one-step conditional law does not explicitly recover the dependence on earlier resolved states.

We have also applied the IEnKF-G and competing DA methods to the more common L96-2s model to the same extent as the GL15
experiments. However, the results are qualitatively very similar to those obtained for GL15, so that we omit them to
avoid repetition.  The model parameters are consolidated in Table~\ref{tab:models_comparison}, along with the
observational parameters that will be used in the DA experiments.  The configurations and the model parameters of GL15\
that are chosen result from a compromise between maintaining the typical chaotic behaviour of a one-dimensional L96
model (Lyapunov spectrum, autocorrelation) and a significant impact of model error on the dynamics forecasts.

\begin{table}[htbp]
  \centering
  \caption{ \label{tab:models_comparison}
    Models' specifications. The Lyapunov time of L63 is for our stochastic L63. The Lyapunov time of the large-scale
    GL15 is estimated using the approximate tangent dynamics in Appendix~\ref{app:gl15}.}
  \begin{tabular}{ccccccc}
    \toprule
    Model & $\Nx$ (full dim.)  & Parameters & Lyap. time & $\Delta_t$ & $\mathbf H$ & $\sigma_\mathrm{obs}$ \\
    \midrule
    L63 & 3 (3) & $s=10$, $r=28$, $b=8/3$, $\sigma=0.25$ & $1.17$ & $0.10$, $0.25$, $0.50$ & $xyz$, $xz$ & $2$ \\
    GL15 & 40 (160) & $J=4$, $h=0.3$, $F=8$ & $1.14$ & $0.05$ & all var. & $1$ \\
    \bottomrule
  \end{tabular}
\end{table}

\subsection{A stochastic Lorenz 1963 model}

The L63 variant we consider incorporates model noise which multiplicatively depends on the state variables
\citep{chekroun2011,savary2026}:
\begin{subequations}
  \begin{align}
    \mathrm{d} x &= s(y - x)\mathrm{d}t + x \sigma \, \mathrm{d} W^x_t, \\
    \mathrm{d} y &= \left[x(r - z) - y\right]\mathrm{d}t  + y \sigma \, \mathrm{d} W^y_t, \\
    \mathrm{d} z &= (xy - bz)\mathrm{d}t + z \sigma \, \mathrm{d} W^z_t,
  \end{align}
\end{subequations}
where $s=10$, $r=28$, $b=8/3$.  Each equation is perturbed by linearly multiplicative noise in the Itô interpretation,
with its own independent Wiener process $W_t^{x,y,z}$, which is scaled with noise amplitude $\sigma=0.25$. Owing to its
state-dependent stochastic model error, the model is integrated using a Milstein scheme with an integration time step of
$\delta_t=10^{-3}$.

\subsection{The Grooms--Lee 2015 model}

The GL15 model evolves a single fine-grid prognostic vector $\bv$ of dimension $J\Nx$, where $\Nx$ is the number of
large-scale grid points and $J$ is the scale-separation factor. The governing equation is
\begin{subequations}
\begin{equation}
\frac{\mathrm{d}\bv}{\mathrm{d}t} = h \NL_{\bv}(\bv) + J \bPi\T \NL_{\bx}(\bPi\bv) - \bv\ + F \bone_{J\Nx},
\end{equation}
where $\bPi$ is a linear projection from the fine grid to the large-scale, coarse grid, $\bPi\T$ is its transpose, $F$
is the forcing scale, and $h$ controls the strength of the small-scale nonlinear dynamics.  The fine-scale nonlinear
operator $\NL_{\bv}$ is defined component-wise by
\begin{equation}
  \left[\NL_{\bv}(\bv)\right]_i = v_{i+1}\left(v_{i-1}-v_{i+2}\right),
  \qquad i=1,\ldots,J\Nx
\end{equation}
with periodic indexing. The large-scale variable is obtained by projection:
\begin{equation}
\bx = \bPi\bv,
\end{equation}
and the large-scale nonlinear operator $\NL_{\bx}$ is
\begin{equation}
\left[\NL_{\bx}(\bx)\right]_i = x_{i-1}\left(x_{i+1}-x_{i-2}\right),
\qquad i=1,\ldots,\Nx,
\end{equation}
\end{subequations}
again with periodic indexing.
Thus the model can be written component-wise as
\begin{equation}
\frac{\mathrm{d}v_i}{\mathrm{d}t} = h\left[\NL_{\bv}(\bv)\right]_i + \left[J \bPi\T \NL_{\bx}(\bPi\bv)\right]_i - v_i + F.
\end{equation}
The large-scale component of $\bv$ on the fine grid is $J \bPi\T \bPi \bv $, and the corresponding
small-scale residual is $\bv - J \bPi\T \bPi \bv$.

Following \citet{grooms2015,grooms2021} we choose for $\bPi$ a projection onto the first $\Nx$ Fourier modes followed by an
inverse Fourier transform to the $\Nx$-dimensional state space.  For even $\Nx$ and $J\ge 2$, we retain the fine-grid
cosine component at the coarse-grid Nyquist wavenumber $\Nx/2$, divide its amplitude by $\sqrt{2}$, and discard the
corresponding sine component. This extends the construction of \citet{grooms2015} to even $\Nx$ while preserving the
normalisation
\begin{equation}
  \bPi\bone_{J\Nx} = \bone_{\Nx},
  \qquad
  J\bPi\bPi\T = \bI_{\Nx}.
\end{equation}

As a result, when $h=0$, the large-scale variable $\bx=\bPi\bv$ obeys the
single-scale L96 equation
\begin{equation}
  \label{eq:gl15-ls}
  \frac{\mathrm{d}\bx}{\mathrm{d}t} = \NL_{\bx}(\bx) - \bx + F\bone_{\Nx}.
\end{equation}
The true dynamics will be defined by the sequence of the large-scale state vectors $\bx\in\R^{\Nx}$.

In the experiments, we choose $\Nx=40$, $J=4$, $h=0.3$, $F=8$. The model is integrated using a fourth-order Runge--Kutta
scheme with an integration time step of $\delta_t = 0.025$. A trajectory of the large-scale model together
with the corresponding fine-scale trajectory (Hovmöller plots) is illustrated in Appendix~\ref{app:gl15}. Estimated
global characteristics of the dynamics are also reported.

\subsection{Training the surrogate models}
\label{sec:training}

The L63 neural network is based on a simple dense architecture, whereas the large-scale GL15 neural network adopts a
simple convolutional architecture. For both dynamics, we have trained probabilistic surrogate models using (i) a
stochastic model trained with an ensemble score-based loss function, (ii) a flow-matching model based on the formulation
of interpolants \citep{lipman2023,chen2024b,albergo2025}. These variants of trained generative surrogate models were
assessed using their forecast skill. The two approaches gave similar forecast root-mean-square errors (RMSEs). In our
tests, flow-matching models trained faster but had slower inference. We therefore use the ensemble score-trained models
for the repeated ensemble forecasts required in our DA experiments.

The surrogates used in the reported DA experiments were trained with the fair energy score based on the Euclidean norm,
as detailed in Appendix~\ref{app:ensemble-score-trained}. In the GL15 case, a training based on the fair CRPS was also
tested. Yet, the large-scale GL15 surrogates trained with the fair CRPS and the fair energy score yielded
indistinguishable forecast performance in the tested configurations. This similarity may reflect the short-range
correlations in GL15 and the spatial inductive bias of the convolutional architecture.  Hence, \emph{ensemble
score-trained} is how we will qualify those surrogates. This is detailed in Appendix~\ref{app:ensemble-score-trained}.

For comparison we also trained deterministic models. One particular deterministic model is the stochastic surrogate but
specialised to the case $\err=\bzero$. We empirically found that its forecast skill is at least as good as that of the
deterministic surrogates.  Nonetheless, if $\Mc$ is nonlinear, $\Mc(\bx_0,\bzero)$ generally departs from
$\Ave{\err}{\Mc(\bx_0,\err)}$ or from the conditional mode at $t_1$. Moreover, in high dimension, $\err=\bzero$ is
atypical, and its neighbourhood is rarely sampled during training, because the Gaussian seed distribution concentrates
near radius $\sqrt{\Nerr}$.

Finally note that the training, validation, test, and DA datasets are mutually independent.

\subsection{Baseline data assimilation methods, hyperparameter tuning, and scores}

The IEnKF-G's performance is compared to three state-of-the-art DA methods.

The EnKF-AdInf method \citep{raanes2019a}, as defined in their Section~5, is a well-tuned EnKF which accounts for both
sampling and model error via an adaptive multiplicative inflation of the prior. As a result, the EnKF-AdInf does not
require any hyperparameter tuning, and is hence very fast to evaluate with little overhead computational cost. The
EnKF-AdInf is representative of the mitigation strategy that accounts for model errors without correcting them.

The IEnKF-Q \citep{sakov2018b} is based on the IEnKF. It handles nonlinearity over one time step $\Delta_t$ like the
IEnKF-G, but assumes that model error is additive and Gaussian.  As explained earlier, the Bayesian formulations of the
IEnKF-Q and IEnKF-G coincide for additive Gaussian model error with state-independent covariance $\bQ$. Their ensemble
implementations may nevertheless differ for nonlinear operators, since they use different ensembles to estimate
sensitivities.  However, it is not adaptive in its basic design. Hence, both the multiplicative prior inflation
$\lambda$ and the magnitude $\sigma_q$ of the additive Gaussian noise are tuned. The selected values of the
hyperparameters are chosen to maximise the filtering accuracy.  As a result of those explicit parameters, the IEnKF-Q is
numerically costly to evaluate. The target model error covariance is $\bQ=\sigma_q^2\bI_{\Nx}$. It is represented by a
centred ensemble of $\Nq$ perturbations (obtained from Algorithm~\ref{alg:regular-simplex}). For the L63 experiments,
$\Nq=\Nx+1=4$, so that the seed space is the full state space.  For the GL15 experiments, $\Nq=\Nx+1=41$, so that
empirical covariance matches the targeted $\bQ$.

Since we will test the simplified SDA IEnKS-G with $L=10$, we choose to use the SDA IEnKS as defined by
\citet{bocquet2014a} as another baseline with a similar $L$-long DAW. With the IEnKS, model error is only mitigated
through multiplicative inflation $\lambda$. The selected $\lambda$ is chosen to maximise the filtering accuracy.  We
need to specify the forecast model used by these three advanced but classical DA methods. In the case of the L63 model,
the EnKF-AdInf, the IEnKF-Q and the IEnKS rely on the exact L63 drift without the stochastic perturbations.  In the case
of the GL15 model, these methods rely on the deterministic model which results from the learned stochastic model by
setting the seed $\err$ to $\bzero$. We found this model to be the most accurate of all the deterministic models we
experimented with. In particular, we checked that it is at least as accurate as a second-order and a third-order
polynomial Wilks statistical correction to the L96 model \citep{wilks2005}.

For the IEnKF-G and IEnKS-G, the retained multiplicative inflation parameter $\lambda$ maximises the filtering accuracy.
However, we found that with our low-order models, this just offers a slight refinement over the case without inflation
(see Table~\ref{tab:methods_comparison}). The exception is the case $\Ng=1$ which amounts to using a deterministic model
and where some small but non-negligible multiplicative inflation is expected ($\lambda \sim 1.05$). In all the
experiments, we report the results of their transform variants.

In all DA methods using multiplicative inflation, the state ensemble anomalies are multiplied by $\lambda$ immediately
before each analysis. The seed anomalies are unchanged.  Table~\ref{tab:methods_comparison} compares the hyperparameters
of the DA methods.

\begin{table}[htbp]
  \centering
  \caption{\label{tab:methods_comparison}
    Comparison of hyperparameters for the DA methods. $\sigma_q$ is the amplitude of Gaussian additive noise, $\lambda$
    is the multiplicative inflation parameter. \textit{opt.} means optimised, \textit{mult. infl.} means multiplicative
    inflation, \textit{add. err.} means additive error.
  }
  \begin{tabular}{c|c|cc|cc}
    \toprule
    Method & DAW & \multicolumn{2}{c|}{Addressing model error} & \multicolumn{2}{c}{Parameter range}   \\
           &            &  Coping mechanism    & Further tuning     &      L63 & GL15                   \\  
    \midrule
    EnKF-AdInf &  & adaptive mult. infl. &  & &  \\
    IEnKF-Q & $[t_0, t_1]$ & opt. Gaussian add. err.  & $\lambda$ & $\lambda\in [1,2.60]$ & $\lambda\in [1, 1.05]$  \\
    \textendash &  &  & $\sigma_q$ & $\sigma_q\in[0, 3.90]$ & $\sigma_q\in [0, 0.05]$  \\
    SDA IEnKS & $[t_0, t_L]$ &  & $\lambda$ & & $\lambda\in [1,1.10]$  \\
    IEnKF-G & $[t_0, t_1]$ & opt. generative seed & $\lambda$  & $\lambda\in [1,1.10]$ & $\lambda\in [1,1.10]$  \\
    SDA IEnKS-G & $[t_0, t_L]$ & opt. generative seed & $\lambda$ &  & $\lambda\in [1,1.10]$  \\
    \bottomrule
  \end{tabular}
\end{table}

The runs of all DA methods are $K=4096$ $\Delta_t$-time steps long after a burn-in of $1024$ time steps.  All DA method
runs are replicated $10$ times except for the IEnKF-Q runs which are replicated $5$ times, each run with a full set of
independent random seeds.  The RMSE is used as the main statistical indicator. Assuming $\bE_k=\left\{ \bx^{(i)}_k
\right\}_{i=1,\ldots,\Ne}$ is the analysis ensemble at $t_k$, $\bar\bx_k$ its mean, and $\xt_k$ the truth at $t_k$, the
RMSE is defined here by
\begin{equation}
  \mathrm{RMSE}(\bar\bx,\xt) = \frac{1}{K}\sum_{k=1}^K \rho\lp\bar\bx_k-\xt_k\rp,
\end{equation}
where $\rho(\bx) = \left\| \bx\right\|_2/\sqrt{\Nx}$ is the coordinate-averaged Euclidean norm. 
The (coordinate-averaged) fair CRPS is also used and defined here by
\begin{subequations}
\begin{equation}
\mathrm{ES}(\bE,\xt) = \frac{1}{K} \sum_{k=1}^K \mathrm{ES}_{\Ne,\rho}(\bE_k,\xt_k),
\end{equation}
where
\begin{equation}
  \label{eq:fair_ensemble_score}
  \mathrm{ES}_{\Ne,\rho}(\bE_k,\xt_k) = 
  \frac{1}{\Ne} \sum_{i=1}^{\Ne} \rho\lp\bx_k^{(i)}-\xt_k\rp
  - \frac{1}{\Ne(\Ne-1)} \sum_{1\leq i<j\leq\Ne}
    \rho\lp\bx_k^{(i)}-\bx_k^{(j)} \rp,
\end{equation}
and where $\rho(\bx) = \left\| \bx\right\|_1/\Nx$ is the coordinate-averaged $L_1$ norm.
\end{subequations}
Because the runs are replicated, both the mean and the standard deviation across the replicated time-averaged scores are
provided.

\subsection{Data assimilation experiments with the Lorenz 1963 model}

The DA experiments with the stochastic L63 model explore the applicability of the IEnKF-G under increasingly nonlinear
conditions. We consider three observation intervals, $\Delta_t\in\left\{0.10, 0.25, 0.50\right\}$, corresponding to
weakly, mildly, and strongly nonlinear regimes, respectively. Two observation configurations are studied: $xyz$, where
all three variables are observed, and $xz$, where only the first and last variables are observed. The observation errors
are all independent, unbiased and Gaussian, with standard deviation $\sigma_\mathrm{obs}=2$. Those values of $\Delta_t$
and $\sigma_\mathrm{obs}$ are taken from the DA literature \citep[e.g.,][]{bocquet2011c}.

\begin{figure*}[t]
  \centering
    \includegraphics[width=0.95\textwidth]{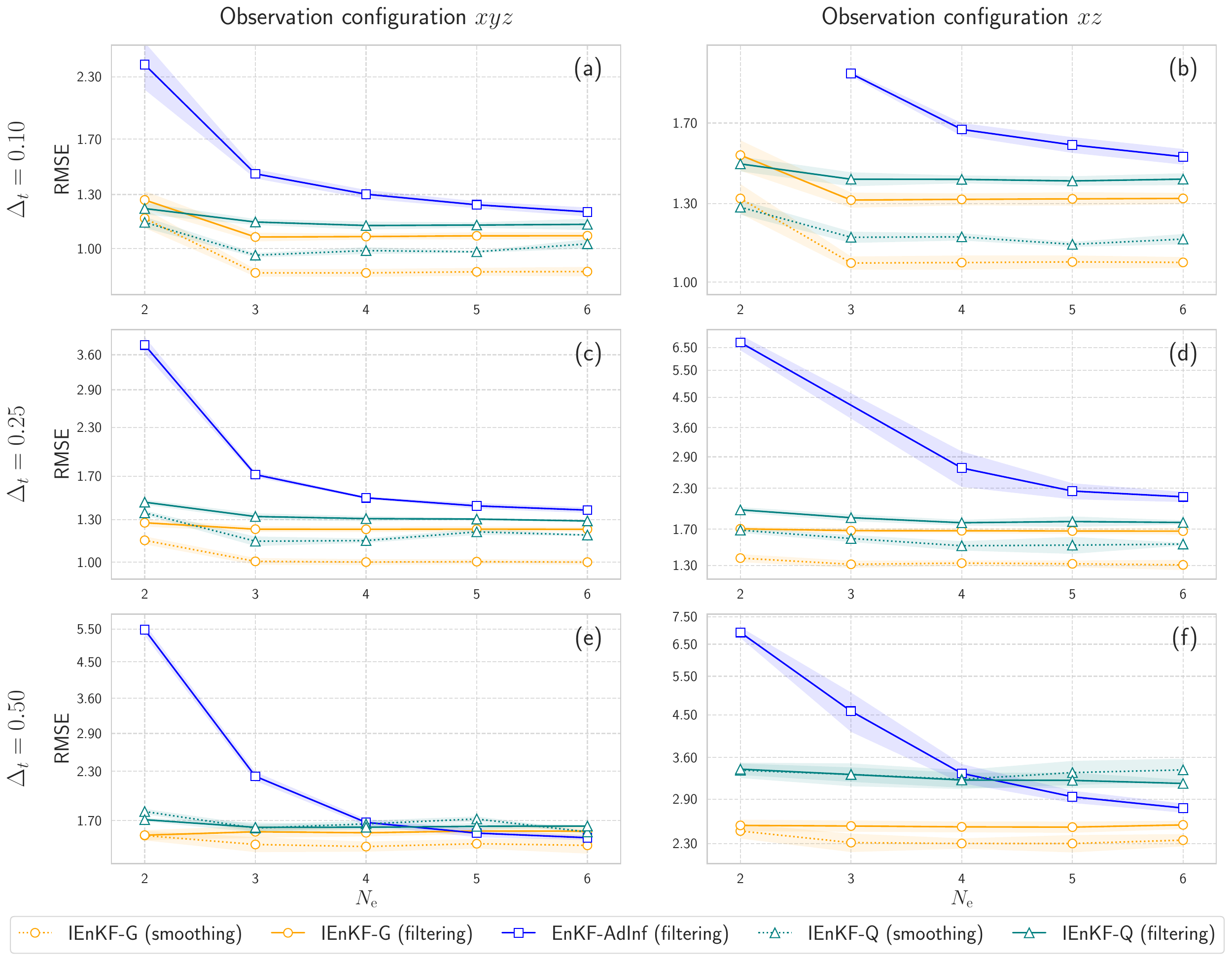} 
  \caption{ \label{fig:l63exp}
    Data assimilation methods comparison using the stochastic L63 model, in two observation configurations ($xyz$, $xz$)
    and three time intervals $\Delta_t$ between observations. Solid and dotted curves denote filtering and one-step
    smoothing RMSEs, respectively. The shaded areas around the curves indicate plus or minus twice the standard deviation across
    replicates. Missing RMSE values indicate divergence of the DA runs.}
\end{figure*}

We choose $\Ng=4$ for the IEnKF-G so as to span the full seed space. The RMSEs are reported in Fig.~\ref{fig:l63exp} as
a function of the ensemble size $\Ne$ ranging from $2$ to $6$.  For the IEnKF-Q, we choose $\Nq=4$, in accordance with
the choice $\Ng=4$. Both the filtering RMSE (at present time) and the smoothing RMSE ($\Delta_t$ in the past) for the
IEnKF-G and the IEnKF-Q are displayed.

Note that in the case $\Delta_t=0.50$ multimodality can often emerge in the cost functions.  As a result, the
Gauss--Newton scheme of the IEnKF-Q and the IEnKF-G may occasionally fail.  That is why we used variants in the
implementation of both the IEnKF-Q and the IEnKF-G where the Gauss--Newton optimisation loop is replaced with a
trust-region method that improves robustness in these significantly nonlinear cases.  We used the trust-region
reflective method implemented in SciPy's \textsf{least\_squares} solver (\textsf{method='trf'}).  This is only helpful
in the $\Delta_t=0.50$ case and it is not leveraged in the GL15 set of experiments.

Figure~\ref{fig:l63exp} shows that the IEnKF-G reaches nearly constant RMSEs from $\Ne=3$, whereas the EnKF-AdInf
remains strongly sensitive to ensemble size, particularly at longer observation intervals. The IEnKF-Q also shows
relatively little sensitivity to $\Ne$. This behaviour is consistent with the control over the seed and the
Gaussian perturbation in both iterative methods: increasing the state ensemble alone is less consequential when
directions of model error are already represented.

For $\Ne\ge3$, the IEnKF-G improves on the IEnKF-Q in both filtering and smoothing across all six configurations, but
the magnitude of the improvement depends strongly on the observation setup. With full observations and
$\Delta_t=0.50$, the filtering scores become comparable, and the EnKF-AdInf is competitive at the largest ensemble
sizes. With partial observations at the same interval, however, the IEnKF-G maintains filtering and smoothing RMSEs
near $2.5$ and $2.3$, compared with approximately $3.2$ and $3.4$ for the IEnKF-Q. This contrast suggests that the
learned transition model is especially valuable when observations leave part of the state to be constrained indirectly
through the dynamics.

The smoothing results further distinguish the methods. The IEnKF-G consistently benefits from the additional
observation batch, whereas the IEnKF-Q obtains little benefit, and sometimes deteriorates, at the longest observation
interval. Thus, the generative formulation improves not only the current-state estimate but also the use of
observations to reconstruct the preceding state.

\subsection{Data assimilation experiments with the Grooms--Lee 2015 model}

The GL15 experiments extend the assessment to a spatially distributed system with $\Nx=40$ resolved variables.

\begin{figure*}[t]
  \centering
  \includegraphics[width=0.85\textwidth]{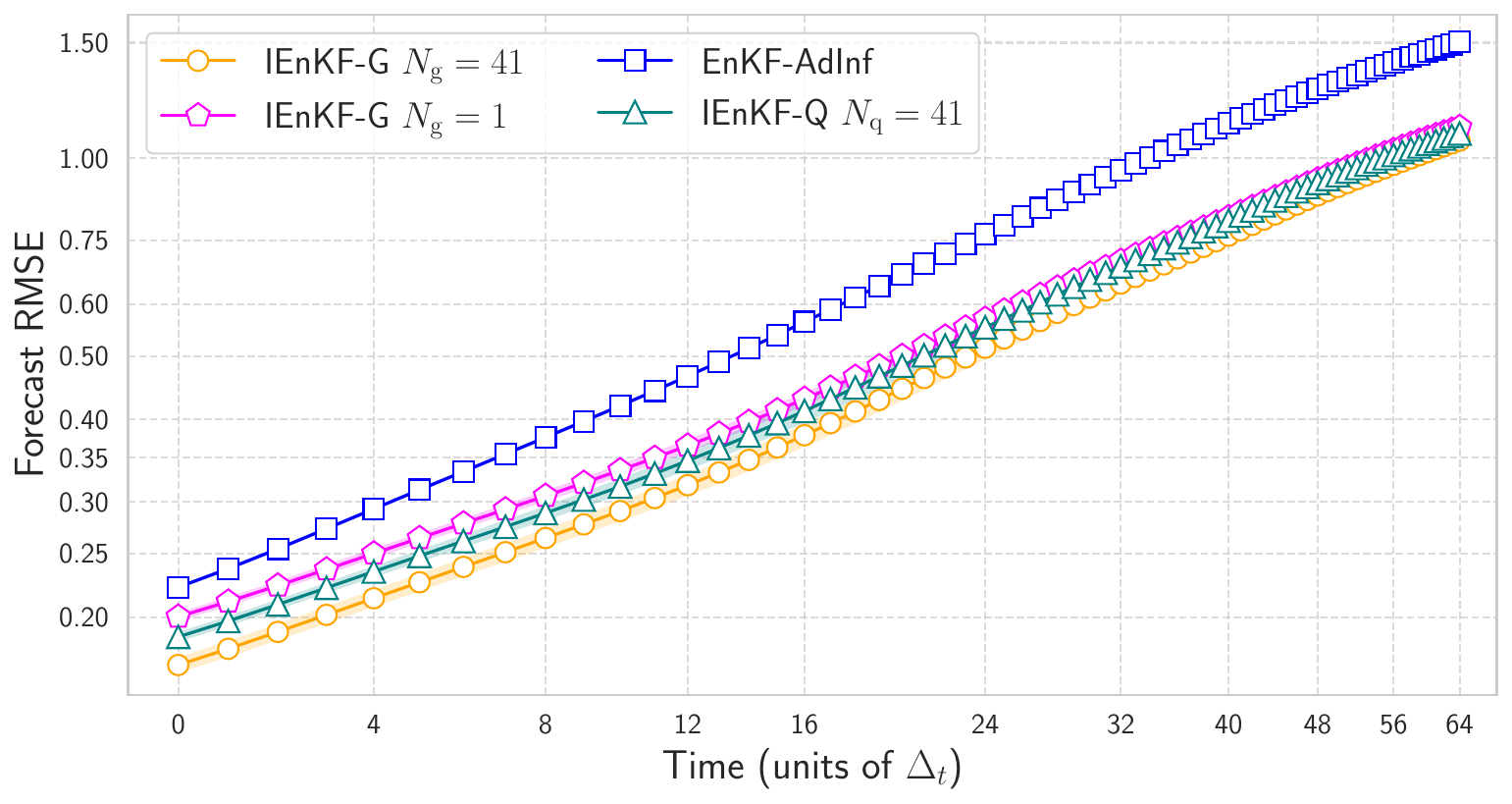} 
  \caption{ \label{fig:gl15-exp1}
    Data assimilation methods comparison using the large-scale GL15 model: forecast RMSEs of the IEnKF-G, IEnKF-Q and
    EnKF-AdInf, as a function of the forecast lead time in units of $\Delta_t$. The shaded areas around the curves
    indicate plus or minus twice the standard deviation across replicates. Note that the RMSEs at time $0$ are the
    filtering RMSEs. }
\end{figure*}

In the first experiment, we compare the IEnKF-G to the baselines when the system is fully observed every $\Delta_t=0.05$
with statistically independent unbiased Gaussian observation errors of standard deviation $\sigma_\mathrm{obs}=1$.  The
ensemble size for the three ensemble DA methods is $\Ne=40$, which avoids the need for localisation.  We compute the
forecast RMSEs as a function of the forecast lead time. They are displayed in Fig.~\ref{fig:gl15-exp1}.  The IEnKF-G is
run with $\Ng=41$, spanning the full seed space, and with $\Ng=1$.  We forecast the ensembles for $64$ time steps
$\Delta_t$ starting from the filtering ensemble and holding the seed to $\bzero$; this corresponds to $\approx 2.81$
Lyapunov times. The filtering ensembles of the EnKF-AdInf are forecasted over the same $64\times\Delta_t$ lead time. The
IEnKF-Q is run with $\Nq=41$, spanning the full perturbation space. Its filtering ensemble is propagated over the same
lead time using the deterministic model with seed $\err=\bzero$.

The IEnKF-G with $\Ng=41$ yields the lowest RMSE throughout the lead times shown in Fig.~\ref{fig:gl15-exp1}, followed
by the IEnKF-Q, the IEnKF-G with $\Ng=1$, and the EnKF-AdInf. At zero lead time, controlling the seed reduces the RMSE
from approximately $0.20$ to $0.17$, and its advantage persists to $64\times\Delta_t$. The IEnKF-Q also improves on the
deterministic configuration of the IEnKF-G, confirming the benefit of explicitly accounting for model error. The further
improvement with $\Ng=41$ supports the use of the learned conditional variability in the analysis.

\begin{figure*}[t]
  \centering
  \begin{tabular}{@{}cc@{}}
    \includegraphics[width=0.48\textwidth]{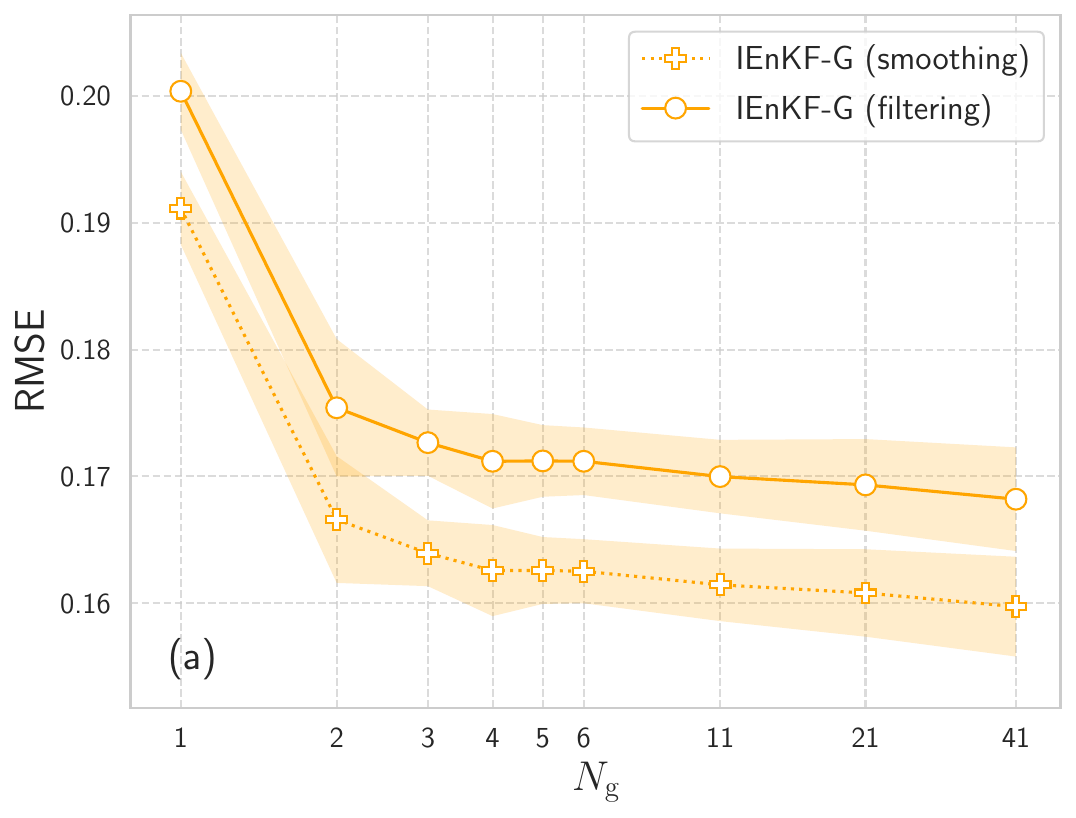} &
    \includegraphics[width=0.48\textwidth]{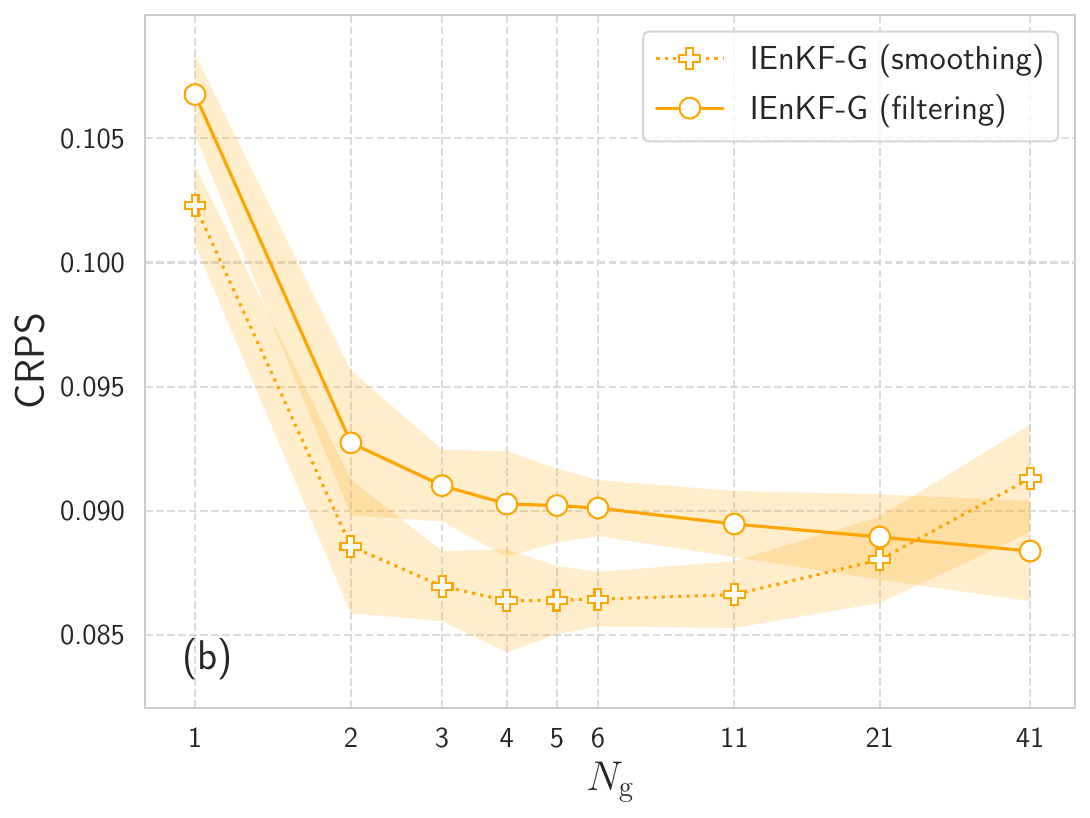}
    \end{tabular}
  \caption{ \label{fig:gl15-exp2}
    Averaged RMSEs (a) and averaged CRPS (b) of the IEnKF-G as a function of $\Ng$, applied to the large-scale GL15
    model. The shaded areas around the curves indicate plus or minus twice the standard deviation across replicates. }
\end{figure*}

In the second experiment, we vary the seed ensemble size $\Ng$ while keeping $\Ne=40$ and the same observation setup.
For $\Ng\ge 2$, this provides $\Ng-1$ independent seed perturbation directions, with the same trace-preserving
normalisation and resampling of their orientation as above. The filtering and one-step smoothing RMSEs are shown in
Fig.~\ref{fig:gl15-exp2}a, and the corresponding ensemble CRPS values in Fig.~\ref{fig:gl15-exp2}b.  The smoothing CRPS
is computed from the $N$-member smoothing ensemble of the IEnKF-G, i.e. using Eq.~\eqref{eq:As0_ienkfg};
Eq.~\eqref{eq:fair_ensemble_score} must be adapted accordingly, in particular $\Ne \rightarrow N$.

Most of the RMSE reduction occurs between $\Ng=1$ and $\Ng=2$, with diminishing gains thereafter for both filtering and
smoothing. Thus, a small seed ensemble already captures much of the benefit of model error control in this
configuration; because its orientation is resampled, this does not imply that a fixed low-dimensional seed subspace
would suffice.  The filtering CRPS follows the same overall improvement. By contrast, the smoothing CRPS reaches its
minimum around $\Ng=4$--$5$ and then increases, eventually exceeding the filtering CRPS at $\Ng=41$, despite
consistently lower smoothing RMSEs.  The smoothing ensemble is underdispersive at $\Ng=41$, with a spread–skill ratio
of approximately $0.77$. This is consistent with its poorer CRPS despite a lower ensemble-mean RMSE. The increasingly
wide seed representation and ensemble approximations may contribute to this difference.

\begin{figure*}[t]
  \centering
  \includegraphics[width=0.85\textwidth]{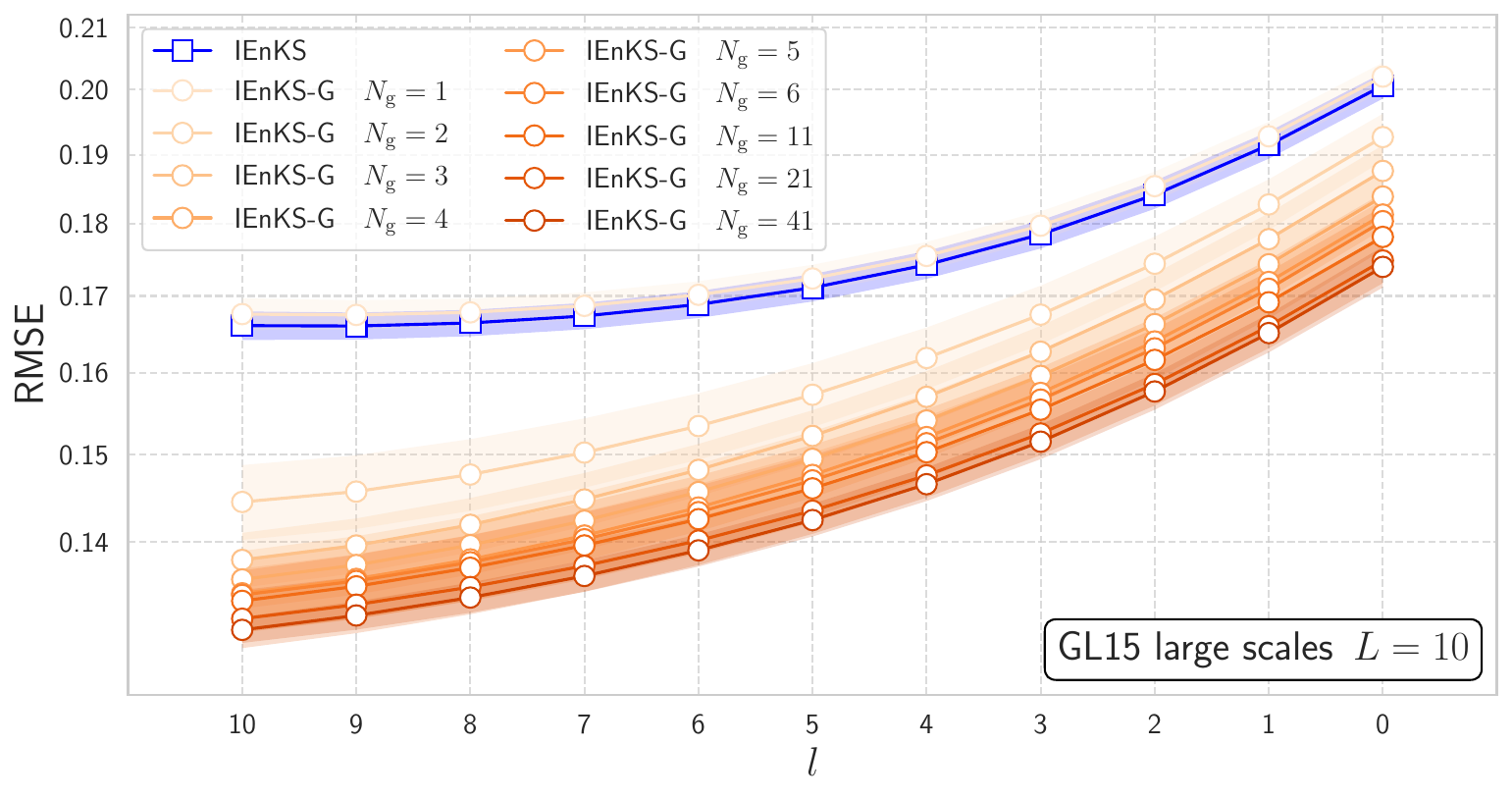}
  \caption{ \label{fig:gl15-exp3}
    Averaged RMSEs of the SDA IEnKS-G and of the SDA IEnKS applied to the large-scale GL15 model, for several values of
    $\Ng$, as a function of the lag $0\le l \le L$ with $L=10$. The shaded areas around the curves indicate plus or
    minus twice the standard deviation across replicates. }
\end{figure*}

In the third experiment, we evaluate the simplified SDA IEnKS-G with $L=10$, $\Ne=40$ and $\Ng$ in the range $1$--$41$.
Figure~\ref{fig:gl15-exp3} reports the RMSE as a function of the lag $0\le l \le L$: $l=0$ denotes filtering, and $l\ge
1$ denotes estimation of the state $l\times\Delta_t$ in the past. Whenever $L\ge 2$, the lag-$l$ state estimates for $0
\le l \le L-2$ are not prescribed by Algorithm~\ref{alg:ienksg-transform}; they are taken as the successive means of the
sensitivity ensemble propagated through the entire DAW (other choices are possible).  Each SDA update assimilates only
the latest observation batch, with earlier observations accounted for through the cycled prior, and uses a single shared
seed within the window. The baseline is the SDA IEnKS with the deterministic surrogate and tuned multiplicative
inflation, but no explicit model error control.

Figure~\ref{fig:gl15-exp3} shows that seed control strengthens the benefit of smoothing. For $\Ng=1$, the IEnKS-G and
the IEnKS perform equally within the error margin. This is expected since both methods coincide in their cost function,
being both deterministic and relying on the surrogate generative model with seed $\err=\bzero$. Yet, again, their
ensemble implementations differ slightly: for $\Ng=1$, the IEnKS-G retains one additional central column in its
sensitivity ensemble used in the Gauss--Newton solver.

For $\Ng\ge 2$, the IEnKS-G outperforms the IEnKS at every displayed lag, with little additional improvement from
$\Ng=5$ to $41$. At $\Ng=41$, the RMSE decreases from approximately $0.175$ at $l=0$ to $0.13$ at $l=10$, about $20\%$
below the IEnKS at the latter lag.  These results show that the shared-seed approximation can improve both filtering and
retrospective state estimation, provided enough seed directions are controlled.  Comparing to the IEnKF-G, i.e. for
$l\in\left\{ 0, 1\right\}$, one notices that the IEnKS-G $\Ng=41$ is less accurate at $l=0$, $l=1$.  This degradation
increases with $L$ (not shown): the shared-seed approximation over the DAW likely costs accuracy at short lags.  For
longer $L$ (not shown), the skill of the IEnKS-G keeps improving at a slow pace, while the IEnKS skill significantly
degrades. The IEnKS-G skill eventually degrades beyond $L=15$ (not shown), presumably due to the loss of numerical
accuracy and the shared-seed approximation.

\section{Discussion and conclusion}
\label{sec:conclusion}

In this paper, we have introduced the generative iterative ensemble Kalman filter (IEnKF-G), which combines a learned
generative transition model with iterative ensemble variational DA. The initial state and the Gaussian seed driving the
surrogate are jointly inferred from observations, allowing for state-dependent, non-Gaussian model uncertainty without
evaluating the transition density.  The IEnKF-G method generalises the IEnKF-Q, while retaining a local Gaussian
approximation of the augmented posterior. We have also proposed a simplified variant of the generative iterative
ensemble Kalman smoother (IEnKS-G), in which a single shared seed is controlled across a multi-step DAW.

The experiments with stochastic L63 and the resolved GL15 dynamics demonstrate the benefit of correcting model error
through the seed. The IEnKF-G generally improves on the IEnKF-Q and the EnKF-AdInf, with substantial gains in partially
observed, strongly nonlinear L63 configurations and sustained forecast improvements for GL15. In the latter, a small
seed ensemble with resampled perturbation directions captures most of the benefit, and the IEnKS-G further improves
retrospective state estimation. However, lower smoothing RMSEs do not always translate into lower CRPS values. These
results encourage further investigation of the ensemble reliability alongside the state estimation accuracy.

Several theoretical extensions could be considered. In a correlated-prior variant, the updated seed would be retained
across cycles, together with its correlations with the state. A single ensemble of state--seed pairs would then avoid
the ensemble reduction required by the IEnKF-G. Such persistence would introduce temporal memory, but also an additional
modelling assumption that is not determined by learning a one-step transition law. Another extension would supplement
the generative forecast with independent additive Gaussian noise of covariance $\bQ$.  This could account for residual
errors not adequately represented by the surrogate. Formally, this noise could be absorbed into an enlarged seed, so
this extension would remain a particular case of the generative framework.  Its interest would lie in separating learned
conditional variability from a prescribed residual error component.

The same variational formulation could also underpin an ensemble of data assimilations \citep[EDA,
  e.g.,][]{bonavita2012}. Each member would solve a weak-constraint 4D-Var problem in which the learned transition
surrogate replaces the additive Gaussian model error representation. The control variables would be the initial state
and the seed, or a sequence of seeds for a multi-step window. Perturbing the observations and the centres of both the
state and seed prior penalties according to their prescribed covariances would generate an ensemble of optimised
trajectories. Automatic differentiation through the surrogate could supply the gradients required by these generalised
weak-constraint 4D-Var problems. The ensemble of solutions would provide a way to estimate analysis uncertainty,
although the ensemble would not generally constitute exact posterior samples for nonlinear dynamics.  Within each EDA
member, a single seed shared across the time steps of the assimilation window could reduce the control dimension, with
the same temporal modelling qualification as for the simplified IEnKS-G.  This scheme would generalise the
bias-correcting weak-constraint 4D-Var currently implemented in the Integrated Forecasting System \citep{laloyaux2020}.

Finally, scaling up the IEnKF-G and IEnKS-G would require controlling sampling errors and computational cost.
Localisation of state covariances is a natural starting point, but the seed representation also deserves attention. Its
prescribed prior covariance is already the identity; localisation would concern its finite-ensemble approximation and
the seed and state--seed correlations induced by assimilation. If seed components have a spatially local influence
through the surrogate, joint local analyses could constrain both state and seed using nearby observations. For a
non-local latent representation, a possible alternative would be to localise the covariances of the propagated seed
perturbations in physical space \citep{malartic2022}. Either way, localisation should treat the state--seed coupling
consistently and, for the smoother, account for propagation across the DAW. Assessing these choices with calibrated
surrogates in higher-dimensional systems is a necessary next step.

\bmsubsection*{Author Contributions}
MB performed the mathematical, algorithmic, and numerical analysis. MB and TF discussed the theory and numerical
results. All the authors worked on the structure of the manuscript. All the authors reviewed and edited the manuscript.

\bmsubsection*{Conflicts of Interest}
The authors declare no conflicts of interest.

\bmsubsection*{Declaration of Generative AI and AI-assisted technologies in the writing process}

No AI was used in the ideas, concepts, mathematical derivations, algorithmic developments. AI was used to find typos and
inconsistencies in the notation. It was used to check the correctness and consistency of the algorithms. It was used to
suggest rephrasing several paragraphs of the main text. The AI tools used for the reviews are GPT-6 and Opus 5.5. After
using these tools, the authors reviewed and edited the content as needed and take full responsibility for the content of
the publication.

\bmsubsection*{Code availability}
The key algorithmic pieces of code will be made publicly available upon acceptance of the manuscript.

\bmsubsection*{Acknowledgments}
The authors are grateful to Alban Farchi for an in-depth review of the manuscript and for his suggestions.  Tobias
S. Finn and Marc Bocquet acknowledge the support of France 2030 PEPR Maths-Vives, grant ANR-23-EXMA-0003, project
Climaths/GenClimEx.  Tobias S. Finn, Sibo Cheng, and Marc Bocquet acknowledge the support of France 2030 PEPR
Maths-Vives, grant ANR-24-EXMA-0002, project DRUIDS.  Sibo Cheng acknowledges the support of the French Agence Nationale
de la Recherche (ANR) under reference ANR-22-CPJ2-0143-01.  This work has also benefited from the project
\textit{Learning efficient   Data Assimilation from Artificial Intelligence (DAbyAI)} supported by NVIDIA and their
Academic Grant Program through the grant of two RTX 6000 Ada GPUs that were intensively used in the numerical
experiments of this work. CEREA is a member of Institut Pierre-Simon Laplace (IPSL).

\bibliography{references}

\appendix

\bmsection{Algorithms \label{app:algorithms}}
\vspace*{12pt}

In the following pseudocodes, the convergence criterion of optimisation loops is set by the tolerance $\epsilon$ which
applies to the non-dimensional $\|\Delta\bu\|$ and is chosen to be $\epsilon= 10^{-3}$. The maximum number of
Gauss--Newton iterations is set to $40$ (not shown in the pseudocodes). The symbols $\bX$ and $\bY$ in these pseudocodes
typically denote normalised anomaly matrices while $\bA$ usually denotes an unnormalised anomaly matrix. The pieces of
pseudocode highlighted in red show changes relative to the IEnKF algorithm in the absence of model error. $\SRA(\bX,
\Ne, \nu)$ (size reduction/augmentation) denotes ensemble size reduction and potentially augmentation from $N=\Ne+\Nq$
or $N=\Ne+\Ng$ to $\Ne$, as described in Appendix~\ref{app:samplers}, Algorithm~\ref{alg:redux-augment-rr}. $\nu$ is a
flag indicating whether the function applies to normalised (true) or unnormalised anomalies (false). $\Ea_0$ is the
ensemble matrix which serves as the state ensemble prior in all those schemes. $\Xq_1$ and $\Xg_0$ are normalised and
centred anomaly matrices of model error or seed perturbations. In particular $\Xq_1\bone_{\Nq} = \bzero$ and
$\Xg_0\bone_{\Ng} = \bzero$. Note that all matrix square roots are symmetric positive-definite square roots.

Throughout the pseudocodes, model and observation operators act columnwise on ensemble matrices. When an operator takes
two ensemble arguments, corresponding columns are paired. For any matrix $\bX$, the notation $[\bX]_{i:j}$ denotes rows
$i$ to $j$, inclusive, with all columns retained. More generally, the vector and matrix indices are one-based and slice
endpoints are inclusive; $:$ denotes all rows or columns. $\langle \cdot \rangle$ denotes the mean over the columns of
the matrix.

For the IEnKF-G and IEnKS-G, analysed ensembles and mean states \emph{within} the DAW can be obtained by propagating the
state components of the joint $N$-member posterior smoothing ensemble, before ensemble size reduction. Each state member
is propagated with its paired diagnosed seed, which is held fixed throughout the DAW. After ensemble size reduction, the
filtering ensembles have $\Ne$ members. Forecasts beyond the DAW are initialised from these filtering ensembles and can
use either the deterministic model with $\err=\bzero$, as in the forecast experiment reported in
Fig.~\ref{fig:gl15-exp1}, or independent Gaussian seeds drawn for each member at each forecast step.

\subsection{Algorithm of the transform IEnKF-Q}

Algorithm~\ref{alg:ienkfq} presents the pseudocode of the transform variant of the IEnKF-Q. It follows Algorithm~1 of
\citet{sakov2018b}, with two modifications.  First, the ensembles used to estimate observation sensitivities include the
current inferred model error increment $\Xq_1\bu_{\Ne+1:N}$ in their centres. This ensures that the sensitivities with
respect to both control blocks are evaluated consistently with the current forecast estimate. This modification has no
effect for linear observation operators, as used in the numerical experiments of \citet{sakov2018b}.  Second, after
termination, the forecast mean is updated using the final control increment and the last secant approximation, without
an additional model propagation. This correction is exact for an affine model; for a nonlinear model, it does not
recompute changes in the forecast mean caused by the updated ensemble spread.


\begin{algorithm}[!htbp]
  \caption{\label{alg:ienkfq} The transform variant of the \name{IEnKF-Q}, adapted from \citet{sakov2018b}.}
  \begin{algorithmic}[1]
    \setlength{\itemsep}{2.5pt}
         \Require  Prior ensemble $\Ea_0\in\Ex^{\Ne}$,
  $\Ne\geq 2$; normalised and centred $\Xq_1\in\R^{\Nx\times\Nq}$,
  $\Nq\geq 2$; observation $\by_1$, positive-definite $\bR$;
  operators $\Mc$, $\Hc$; tolerance $\epsilon>0$.
\Ensure Analysis ensemble $\Ea_1\in\Ex^{\Ne}$.
    \Function{$[\Ea_1]$ = \textbf{ienkfq\_cycle}}{$\Ea_0{\color{red}, \, \Xq_1}, \, \by_1,\ \bR, \, \Mc, \Hc,
      \epsilon$}
    \State ${\color{red} N=\Ne+\Nq}, \quad \se =\sqrt{\Ne-1}, \quad \sq = \sqrt{\Nq-1}$
    \State $\xa_0 = \langle \Ea_0 \rangle, \quad \Xa_0 = (\Ea_0 - \xa_0 \bone_{\Ne}\T)/\se$
    \State $\bD = \bI_N, \quad \bu = \bzero_N$
    \Repeat
    \State $\bx_0 = \xa_0 + \Xa_0 \bu_{\color{red} 1 : \Ne}$
    \State $\bT = (\bD_{\color{red} 1:\Ne,1:\Ne})^{1/2}$
    \State $\bE_0 = \bx_0 \bone_{\Ne}\T + \se\Xa_0\bT$
    \State $\bE_1 = \Mc (\bE_0) {\color{red} + \Xq_1 \bu_{\Ne+1:N}\bone_{\Ne}\T}$
    \State $\bx_1 = \langle \bE_1 \rangle $
    \State $\bY_1 =  \left[ \Hc(\bE_1) - \langle \Hc(\bE_1) \rangle \bone_{\Ne}\T \right] \bT^{-1}/\se$
    \State ${\color{red} \bE_1^\mathrm{q} = \bx_1\bone_{\Nq}\T +\sq \Xq_1}$
    \State ${\color{red} \bY_1^\mathrm{q} = \left[ \Hc\lp\bE_1^\mathrm{q}\rp
        - \langle \Hc\lp\bE_1^\mathrm{q}\rp \rangle \bone_{\Nq}\T \right]/\sq}$
    \State $\bY = \begin{bmatrix} \bY_1 & {\color{red}\bY_1^\mathrm{q}} \end{bmatrix}$
    \State $\nabla_{\bu}\Lc =  \bu - \bY\T \bR^{-1} [\by_1 - \Hc(\bx_1)]$
    \State $\bD = [\bI_N + \bY\T \bR^{-1} \bY]^{-1}$
    \State $\Delta \bu = -\bD \, \nabla_{\bu} \Lc$
    \State $\bu \gets \bu + \Delta \bu$
    \Until $\| \Delta \bu \| < \epsilon$
    \State $\bX_1 = \left[\bE_1-\langle\bE_1\rangle\bone_{\Ne}\T\right] \bT^{-1}/\se$
    \State $\bx_1 \gets \bx_1 + \bX_1(\Delta\bu)_{1:\Ne} + {\color{red} \Xq_1(\Delta\bu)_{\Ne+1:N}}$
    \State ${\color{red} \bX = \begin{bmatrix}\bX_1 & \Xq_1\end{bmatrix}\bD^\frac{1}{2}}$
    \State ${\color{red} \bX_1 = \SRA(\bX,\Ne,\mathrm{true})}$
    \State $\Ea_1 = \bx_1\bone_{\Ne}\T+\se\bX_1$
    \State \Return $\Ea_1$
    \EndFunction
  \end{algorithmic}
\end{algorithm}

\subsection{Algorithm of the transform IEnKF-G}


Algorithm~\ref{alg:ienkfg-transform} gives the pseudocode of the transform variant of the IEnKF-G. Let
$\bSig=\diag(a_{\mathrm e}\bI_{\Ne},a_{\mathrm g}\bI_{\Ng})$, where $a_{\mathrm e}>0$, and $a_{\mathrm g}>0$ for $\Ng\ge
2$; for $\Ng=1$, allow $a_{\mathrm g}\geq0$. In our experiments, we choose $a_{\mathrm e}=\sqrt{\Ne-1}$, $a_{\mathrm
  g}=\sqrt{\Ng-1}$ but identity scaling is also admissible. With $a_{\mathrm e}=\sqrt{\Ne-1}$, the state perturbations
keep the spread of the prior ensemble, as in the standard transform IEnKF. With $a_{\mathrm g}=\sqrt{\Ng-1}$, the seed
members fed to the surrogate have Euclidean norms comparable to those of the training seeds drawn from $\gauss(0,
\bI_{\Nerr})$, which concentrate around $\sqrt{\Nerr}$.  Define the block-centring projection and the pseudo-inverse by
\begin{equation}
\bC_m=\bI_m-\frac{\bone_m\bone_m\T}{m},\qquad \bP=\diag(\bC_{\Ne},\bC_{\Ng}),\qquad
\bSig^\dagger=\diag(a_{\mathrm e}^{-1}\bI_{\Ne}, a_{\mathrm g}^\dagger\bI_{\Ng}),
\end{equation}
where $a_{\mathrm g}^\dagger=1/a_{\mathrm g}$ if $a_{\mathrm g}>0$, and $a_{\mathrm g}^\dagger=0$ otherwise. The
projection $\bP$ enforces zero sensitivities along the redundant block-constant control directions and preserves
ensemble centring under unequal block scalings. It leaves the sensitivities unchanged for affine operators but modifies
their finite-ensemble approximation for nonlinear operators.

\begin{algorithm}[!htbp]
\caption{\label{alg:ienkfg-transform} The transform variant of the \name{IEnKF-G}.}
\begin{algorithmic}[1]
  \setlength{\itemsep}{2.5pt}
\Require Prior ensemble $\Ea_0\in\Ex^{\Ne}$, $\Ne\geq2$; normalised and centred $\Xg_0\in\mathbb
R^{\Nerr\times\Ng}$, $\Ng\geq1$; observation $\by_1$, positive-definite $\bR$; operators $\Mc$, $\Hc$;
tolerance $\epsilon>0$; block scaling $\bSig$ as defined above.
\Ensure Analysis ensemble $\Ea_1\in\Ex^{\Ne}$.
\Function{$[\Ea_1]=$ \textbf{ienkfg\_transform\_cycle}}{$\Ea_0,{\color{red} \Xg_0},\by_1,\bR,\Mc,\Hc,{\color{red} \bSig},\epsilon$}
  \State ${\color{red} N=\Ne+\Ng}$, \quad $\se=\sqrt{\Ne-1}$
  \State {\color{red} $\bP=\diag(\bC_{\Ne},\bC_{\Ng})$; compute $\bSig^\dagger$ as above.}
    \State $\xa_0 = \langle \Ea_0 \rangle, \quad \Xa_0 = \lp\Ea_0 - \, \xa_0 \bone_{\Ne}\T\rp/\se$
    \State ${\color{red} \za_0 =
      \begin{bmatrix} \xa_0 \\ \bzero \end{bmatrix}, \quad \bX  = \begin{bmatrix} \Xa_0 & \bzero \\ \bzero &
        \Xg_0 \end{bmatrix}}$
  \State $\bD=\bI_N$, \quad $\bu=\bzero_N$
  \Repeat
    \State $\bz_0=\za_0+\bX\bu$
    \State $\bT=\bD^\frac{1}{2}$
    \State $\bZ_0=\bz_0\bone_N\T+\bX{\color{red} \bSig}\bT$
    \State $\bE_1=\Mc\left( {\color{red} [\bZ_0]_{1:\Nx}, [\bZ_0]_{\Nx+1:\Nx+\Nerr}}\right)$
    \State ${\color{red} \dK=\bT^{-1}\bSig^\dagger\bP}$
    \State $\bY=\left[\Hc(\bE_1)
      -\langle\Hc(\bE_1)\rangle\bone_N\T\right]{\color{red} \dK}$
    \State $\bx_1=\langle\bE_1\rangle$
    \State $\nabla_{\bu}\Lc
      =\bu-\bY\T\bR^{-1}\left[\by_1-\Hc(\bx_1)\right]$
    \State $\bD=\left[\bI_N+\bY\T\bR^{-1}\bY\right]^{-1}$
    \State $\Delta\bu=-\bD\nabla_{\bu}\Lc$
    \State $\bu\gets\bu+\Delta\bu$
  \Until{$\|\Delta\bu\|<\epsilon$}
  \State $\bX_1=(\bE_1-\bx_1\bone_N\T){\color{red} \dK\bD^\frac{1}{2}}$
  \State ${\color{red}\bX_1=\SRA(\bX_1,\Ne,\mathrm{true})}$
  \State $\Ea_1=\bx_1\bone_{\Ne}\T+\se\bX_1$
  \State \Return $\Ea_1$
\EndFunction
\end{algorithmic}
\end{algorithm}

\subsection{Algorithm of the bundle IEnKF-G}


Algorithm~\ref{alg:ienkfg-bundle} gives the pseudocode of the $h$--bundle variant of the IEnKF-G, where $h\approx
10^{-4}$--$10^{-2}$ is the finite-difference parameter. Clearly, $h$ can be absorbed into $\bSig$, so that only the
choice of $h\bSig$ matters.

\begin{algorithm}[!htbp]
  \caption{\label{alg:ienkfg-bundle} The bundle variant of the \name{IEnKF-G}.}
  \begin{algorithmic}[1]
    \setlength{\itemsep}{2.5pt}
      \Require  Prior ensemble $\Ea_0\in\Ex^{\Ne}$,
  $\Ne\geq2$; normalised and centred $\Xg_0\in\Ee^{\Ng}$,
  $\Ng\geq1$; observation $\by_1$, positive-definite $\bR$;
  operators $\Mc$, $\Hc$; finite-difference step $h>0$; tolerance $\epsilon>0$; block scaling $\bSig$ as defined above.
\Ensure Analysis ensemble $\Ea_1\in\Ex^{\Ne}$.
    \Function{$[\Ea_1]$ = \textbf{ienkfg\_bundle\_cycle}}{$\Ea_0{\color{red}, \, \Xg_0}, \, \by_1,\ \bR, \,
      \Mc, \Hc, {\color{red} \bSig}, h, \epsilon$}
    \State $ { \color{red} N=\Ne+\Ng,} \quad { \color{red} \sn=\sqrt{N-1},} \quad \se  =\sqrt{\Ne-1} $
  \State {\color{red} $\bP=\diag(\bC_{\Ne},\bC_{\Ng})$; compute $\bSig^\dagger$ as above.}
    \State $\xa_0 = \langle \Ea_0 \rangle, \quad \Xa_0 = \lp\Ea_0 - \, \xa_0 \bone_{\Ne}\T\rp/\se$
    \State ${\color{red} \za_0 =
      \begin{bmatrix} \xa_0 \\ \bzero \end{bmatrix}, \quad \bX  = \begin{bmatrix} \Xa_0 & \bzero \\ \bzero & \Xg_0 \end{bmatrix}}$
    \State  $\bu = \bzero_N$
    \Repeat
    \State $\bz_0 = \za_0 + \bX \bu$
     \State $\bZ_0 = \bz_0 \bone_N\T + h\bX{\color{red} \bSig}$
    \State $\bE_1 = \Mc \left({\color{red} [\bZ_0]_{1:\Nx}, [\bZ_0]_{\Nx+1:\Nx+\Nerr}} \right)$
     \State $\bY = \left[ \Hc(\bE_1) - \langle \Hc(\bE_1) \rangle \bone_N\T \right]{\color{red} \bSig^\dagger\bP}/h$
    \State $\bx_1 = \langle \bE_1 \rangle $
    \State $\nabla_{\bu} \Lc =  \bu - \bY\T \bR^{-1} [\by_1 - \Hc(\bx_1)]$
    \State $\bD = [\bI_N + \bY\T \bR^{-1} \bY]^{-1}$
    \State $\Delta \bu = -\bD \, \nabla_{\bu} \Lc$
    \State $\bu \gets \bu + \Delta \bu$
    \Until $\| \Delta \bu \| < \epsilon$
    \State $\bz_0 = \za_0 + \bX\bu$
    \State $\bZ_0 = \bz_0 \bone_N\T + \sn\bX\bD^\frac{1}{2}$
    \State $\bE_1 = \Mc \left({\color{red} [\bZ_0]_{1:\Nx}, [\bZ_0]_{\Nx+1:\Nx+\Nerr}} \right)$
     \State $\bx_1 = \langle \bE_1 \rangle, \quad \bA_1 = \bE_1 - \bx_1\bone_N\T$
    \State ${\color{red}\bA_1 = \SRA(\bA_1, \Ne, \mathrm{false})}$
    \State $\Ea_1 = \bx_1 \bone_{\Ne}\T + \bA_1$
      \State \Return $\Ea_1$
    \EndFunction
  \end{algorithmic}
\end{algorithm}

\subsection{Algorithm of the simplified transform SDA IEnKS-G}


Algorithm~\ref{alg:ienksg-transform} gives the pseudocode of the transform variant of the simplified SDA IEnKS-G,
i.e. in the case where $S=1 \le L$. The algorithm assimilates the observation $\by_L$ at the end of the window and
returns an ensemble at $t_1$, incorporating this observation together with the information already contained in the
prior ensemble. For $L \ge 2$, this is a smoothed ensemble, which is used to initialise the next window after shifting
it by $S=1$ time step. For $L=1$, it is a filtering analysis ensemble. Notation-wise, $\Mc_{1:L}$ performs $L-1$
forecast steps, reusing the same seed matrix, with $\Mc_{1:1}$ the identity.

\begin{algorithm}[!htbp]
  \caption{\label{alg:ienksg-transform} The transform variant of the simplified SDA \name{IEnKS-G} with $S=1 \le L$.}
  \begin{algorithmic}[1]
    \setlength{\itemsep}{2.5pt}
      \Require  Prior ensemble $\Ea_0\in\Ex^{\Ne}$,
  $\Ne\geq2$; normalised and centred $\Xg_0\in\Ee^{\Ng}$,
  $\Ng\geq1$; observation $\by_L$, positive-definite $\bR_L$;
  operators $\Mc$, $\Hc_L$; DAW length $L\ge 1$; tolerance $\epsilon>0$; block scaling $\bSig$ as defined above.
  \Ensure Ensemble $\Ea_1\in\Ex^{\Ne}$ at $t_1$, updated using $\by_L$.
    \Function{$[\Ea_1]$ = \textbf{ienksg\_cycle}}{$\Ea_0{\color{red}, \, \Xg_0}, \, \by_L,\ \bR_L, \,
      \Mc, \Hc_L, L, {\color{red} \bSig}, \epsilon$}
    \State ${ \color{red} N=\Ne+\Ng,} \quad \se =\sqrt{\Ne-1}$
    \State {\color{red} $\bP=\diag(\bC_{\Ne},\bC_{\Ng})$; compute $\bSig^\dagger$ as above.}
    \State $\xa_0 = \langle \Ea_0 \rangle, \quad \Xa_0 = \lp\Ea_0 - \, \xa_0 \bone_{\Ne}\T\rp/\se$
    \State ${\color{red} \za_0 =
      \begin{bmatrix} \xa_0 \\ \bzero \end{bmatrix}, \quad \bX  = \begin{bmatrix} \Xa_0 & \bzero \\ \bzero &
        \Xg_0 \end{bmatrix}}$
    \State $\bD = \bI_N, \quad \bu = \bzero_N$
    \Repeat
    \State $\bz_0 = \za_0 + \bX \bu$
    \State $\bT = {\bD}^{1/2}$
    \State $\bZ_0 = \bz_0 \bone_N\T + \bX {\color{red} \bSig}\bT$
    \State $\bE_1 = \Mc \left({\color{red} [\bZ_0]_{1:\Nx}, [\bZ_0]_{\Nx+1:\Nx+\Nerr}} \right)$
    \State ${\color{red} \dK=\bT^{-1}\bSig^\dagger\bP}$
    \State $\bE_L = \Mc_{1:L} \left({\color{red} \bE_1, [\bZ_0]_{\Nx+1:\Nx+\Nerr}} \right)$
    \State $\bY =  \left[ \Hc_L(\bE_L) - \langle \Hc_L(\bE_L) \rangle \bone_N\T \right]  {\color{red} \dK}$
    \State $\bx_L = \langle \bE_L \rangle $
    \State $\nabla_{\bu} \Lc =  \bu - \bY\T \bR_L^{-1} [\by_L - \Hc_L(\bx_L)]$
    \State $\bD = [\bI_N + \bY\T \bR_L^{-1} \bY]^{-1}$
    \State $\Delta \bu = -\bD \, \nabla_{\bu} \Lc$
    \State $\bu \gets \bu + \Delta \bu$
    \Until $\| \Delta \bu \| < \epsilon$
    \State $\bx_1 = \langle \bE_1 \rangle $
    \State $\bX_1 =  \lp \bE_1 - \bx_1\bone_N\T \rp {\color{red} \dK\bD^\frac{1}{2}}$
    \State ${\color{red}\bX_1 = \SRA(\bX_1, \Ne, \mathrm{true})}$
    \State $\Ea_1 = \bx_1 \bone_{\Ne}\T + \se\bX_1$
    \State \Return $\Ea_1$
    \EndFunction
  \end{algorithmic}
\end{algorithm}

\bmsection{Ensemble score-trained stochastic surrogate models \label{app:ensemble-score-trained}}
\vspace*{12pt}

\subsection{Conditional transition model}

For each model, the surrogate approximates the one-step conditional law of the current state rather than a deterministic
flow map.  Let $\bx_k\in\Ex$ denote the current state and $\innov_k=\bx_{k+1}-\bx_k$ its increment over the surrogate
time step $\Delta_t$. An ensemble member indexed by $i$ is generated as
\begin{equation}
  \widehat{\bx}_{k+1}^{(i)}
  = \bx_k + \mu_{\delta}\bone_{\Nx}
  + \sigma_{\delta}
    \ssm\!\left(\frac{\bx_k-\mu_x\bone_{\Nx}}{\sigma_x}, \err_k^{(i)} \right),
  \qquad
  \err_k^{(i)}\sim\gauss (\bzero,\bI_{\Nx}),
  \label{eq:surrogate-transition}
\end{equation}
where $\ssm$ is a neural network and the four scalar normalisation statistics $\mu_\delta$, $\sigma_\delta$, $\mu_x$ and
$\sigma_x$ are estimated from the training dataset.

To sample trajectories from the learned transition model, independent seeds $\err_k^{(i)}$ are drawn for each member and
at each $t_k$. Repeated draws at a fixed $\bx_k$ instead provide samples from its one-step transition kernel. The
IEnKS-G, however, uses different temporal seed assumptions. Note that for GL15, this construction provides a Markov
approximation to the projected reference dynamics: even an exact match to their one-step conditional law would not, in
general, reproduce their full temporal dependence.

The network architecture is adapted to the geometry of each state space (Table~\ref{tab:surrogate-architectures}). L63
uses a fully connected network: the normalised three-component state and a three-component Gaussian vector are
flattened, projected to width $8$, passed through $16$ residual blocks, and mapped to a three-component increment. Each
block comprises $2$ affine layers of width $8$, layer normalisation after each layer, a Mish activation after the first
layer, and a skip connection followed by a second Mish activation. A distinct surrogate is trained for each of the three
time steps $\Delta_t$.

GL15 is defined on a periodic one-dimensional grid and therefore uses translation-equivariant circular convolutions. Its
two input channels are the normalised state and the Gaussian seed. A width-$5$ input convolution lifts these channels to
$\Nx$ feature channels; $8$ residual blocks then apply two width-$3$ circular convolutions, with batch normalisation and
Mish activations, before a pointwise output convolution returns the increment. Thus, the same local rule is applied at
every grid point and the periodic boundary condition is built into the network. The residual constructions follow the
general principle of \citet{he2016}.

\begin{table}[htbp]
  \centering
  \small
  \caption{
    Architectures used for the stochastic transition model.  Here $\Netr$ is the number of ensemble members generated
    per training pair, $k_{\mathrm{in}}$ and $k_{\mathrm{b}}$ are the input and residual-block kernel widths,
    respectively, and $N_{\mathrm{tp}}$ denotes the number of trainable parameters.
  }
  \label{tab:surrogate-architectures}
  \begin{tabular}{@{}lccccclr@{}}
    \toprule
    Original model
    & $\Nx$ & $\Delta_t$ & $\Netr$ & Blocks & Channels & Core & $N_{\mathrm{tp}}$ \\
    \midrule
    L63 & $3$ & $0.10$, $0.25$, $0.50$ & $3$ & $16$ & $8$ & dense, $6\!\to\!8\!\to\!3$ & $2899$ \\
    GL15 & $40$ & $0.05$ & $8$ & $8$ & $40$ & conv., $k_{\mathrm{in}}=5$, $k_{\mathrm{b}}=3$ & $79\,201$ \\
    \bottomrule
  \end{tabular}
\end{table}

\subsection{Training data and objective}

Training samples are pairs $(\bx_k,\bx_{k+1})$ produced by the corresponding reference dynamics after a spin-up of
$2048$ time steps $\Delta_t$. L63 data contain the complete three-dimensional state of the stochastic L63 system with
componentwise multiplicative noise of amplitude $\sigma=0.25$. For GL15, the $160$-dimensional high-resolution field is
spectrally projected onto its $\Nx=40$ resolved large-scale variables. In each case $1024$ independent trajectories are
used for the training dataset and another $1024$ but $5$ times shorter are used for the validation dataset.  Successive
initial states are thinned by a factor of $6$, although every target remains one $\Delta_t$ ahead. This gives $524\,288$
training and $104\,448$ validation pairs for L63, and $262\,144$ training and $52\,224$ validation pairs for GL15. The
state and increment are standardised separately using global scalar means and standard deviations computed from the
training pairs.

For each training pair $(\bx_k,\bx_{k+1})$, we define the normalised target increment as
\begin{equation*}
  \widetilde{\innov}_k
  = \frac{\innov_k-\mu_\delta\bone_{\Nx}}{\sigma_\delta}.
\end{equation*}
We sample $\Netr$ independent Gaussian seed vectors and obtain the corresponding normalised increment
forecasts
\begin{equation*}
  \widehat{\widetilde{\innov}}_k^{(i)}
  = \ssm\!\left(\frac{\bx_k-\mu_x\bone_{\Nx}}{\sigma_x}, \err_k^{(i)} \right),
  \qquad i=1,\ldots,\Netr.
\end{equation*}
The training ensemble size $\Netr$ is distinct from the DA ensemble size $\Ne$ and need not have the same
value.

The network parameters minimise the (coordinate-averaged) fair finite-ensemble score:
\begin{equation}
  \label{eq:fair-energy-score}
  \widehat{\mathrm{ES}}_{\Netr,\rho}
  = \frac{1}{\Netr} \sum_{i=1}^{\Netr}
    \rho\left( \widehat{\widetilde{\innov}}_k^{(i)}
      -\widetilde{\innov}_k \right)
  - \frac{1}{ \Netr(\Netr-1)} \sum_{1\leq i<j\leq\Netr}
    \rho\left(\widehat{\widetilde{\innov}}_k^{(i)}
      -\widehat{\widetilde{\innov}}_k^{(j)} \right),
\end{equation}
averaged over the mini-batch.  The first term measures agreement with the normalised target increment, whereas the
second rewards an appropriate ensemble spread; their proper-score balance penalises both collapse and excessive
dispersion \citep{gneiting2007,ferro2014}.  The models used for the results reported in the experiments use
$\rho(\boldsymbol{v})=\|\boldsymbol{v}\|_2/\sqrt{\Nx}$. Yet, in the GL15 case,
$\rho(\boldsymbol{v})=\|\boldsymbol{v}\|_1/\Nx$ was also tested, for which Eq.~\eqref{eq:fair-energy-score} is the
(coordinate-averaged) fair CRPS. However we did not find any meaningful departure between surrogates trained with both
ensemble scores, including in their power spectral density.

All networks are trained with mini-batches of $2048$ examples, freshly sampled Gaussian seeds, validation-based
checkpointing and early stopping. GL15 uses Adamax with an initial learning rate of $10^{-3}$; the rate is multiplied by
$0.8$ after $80$ validation epochs without a relative improvement of $10^{-3}$, down to $10^{-8}$, and training stops
after $240$ unimproved epochs. L63 uses AdamW with learning rate $10^{-3}$, weight decay $10^{-2}$, gradient-norm
clipping at one, a $2\%$ linear warm-up followed by cosine decay to $10^{-8}$, and a patience of $512$ epochs. In all
training experiments, the checkpoint attaining the lowest validation score is used.

\bmsection{Supplements on the GL15 model \label{app:gl15}}
\vspace*{12pt}

\begin{figure*}[!htbp]
  \centering
  \includegraphics[width=0.85\textwidth]{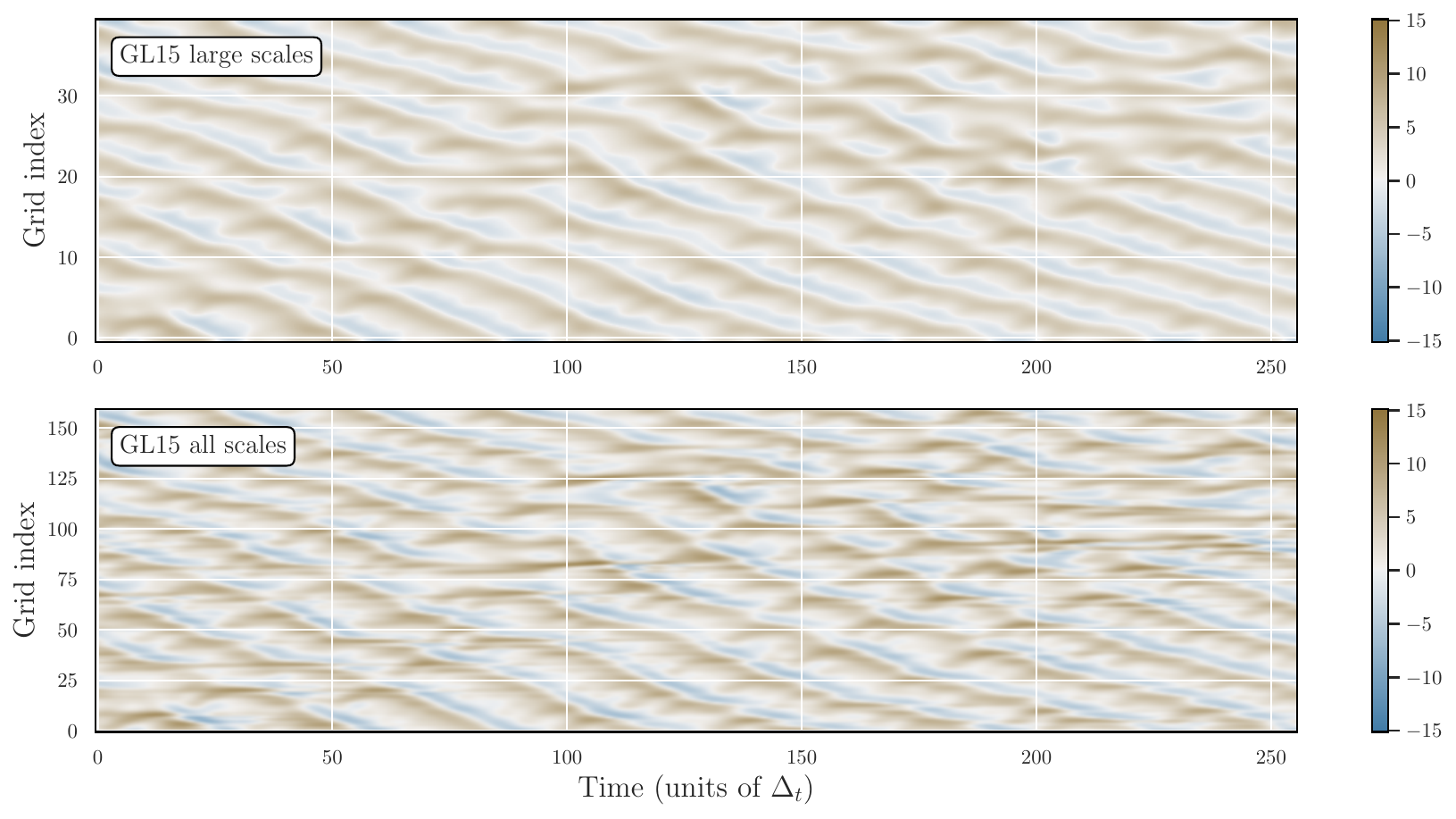} 
  \caption{ \label{fig:gl15-hovmoeller}
    Hovmöller plots of the large-scale variables of a GL15 trajectory (top) and the fine-scale field (bottom) from
    which they are obtained by projection.}
\end{figure*}

Figure~\ref{fig:gl15-hovmoeller} shows a trajectory of the GL15 model, along with the corresponding trajectory of its
large-scale variables.
Moreover, we have assessed the Lyapunov spectrum of the large-scale GL15 dynamics.  To that end, we have computed a
(high-resolution $J\times\Nx=160$) GL15 trajectory along with its (low-resolution $\Nx=40$) large-scale variables
trajectory.

With the approximation that $h=0$, the tangent linear dynamics projected onto the large-scale variables is the tangent
linear of the large-scale dynamics with $h=0$, i.e. the tangent linear of Eq.~\eqref{eq:gl15-ls}. We compute a Lyapunov
spectrum from this approximate tangent linear dynamics, driven by the large-scale variables of the reference trajectory.
This is equivalent to computing the Lyapunov spectrum of the large scale dynamics under the assumption that the
fine-scale feedback onto the large scales is held fixed. Its largest Lyapunov exponent is $0.88$ corresponding to a
Lyapunov time of $\approx 1.14$. There are $9$ positive exponents and one near-zero exponent. The approximate
Kaplan--Yorke dimension is $\approx 20.2$.

\bmsection{Perturbation sampling routines \label{app:samplers}}
\vspace*{12pt}

\subsection{Generating simplex ensembles}

Algorithm~\ref{alg:regular-simplex} constructs a centred ensemble of $\Nout$ vectors in $\R^{\Nx}$, stored as the
columns of $\Zout$. For $\Nout\geq2$, it first obtains an orthonormal basis $\bOmega$ of the subspace perpendicular to
$\bone_{\Nout}$ in $\R^{\Nout}$ by a reduced QR decomposition. The columns of $\bX=\sqrt{\Nout/(\Nout-1)}\,\bOmega\T$
are the vertices of a regular simplex: each has unit norm and any two distinct columns have inner product
$-1/(\Nout-1)$. If $\Nx\geq \Nout-1$, a second QR decomposition, applied to a Gaussian matrix, embeds this simplex
isometrically into a random $(\Nout-1)$-dimensional subspace of $\R^{\Nx}$. Otherwise, the simplex is orthogonally
projected onto a random $\Nx$-dimensional subspace of $\R^{\Nout-1}$ and expressed in an orthonormal basis of that
subspace; equal vertex norms and pairwise distances are then generally lost. The final scaling sets the trace of the
unbiased empirical covariance $\bP=\Zout\Zout\T/(\Nout-1)$ to $\Nx$. The output is centred and has rank
$m=\min(\Nout-1,\Nx)$, with all $m$ nonzero covariance eigenvalues equal to $\Nx/m$, so it is isotropic on its
span. Whenever $\Nout-1\geq\Nx$, $\bP=\bI_{\Nx}$ in exact arithmetic. These properties hold almost surely for the
Gaussian draws and up to roundoff in computation. For $\Nout=1$, the function instead returns a single zero vector. When
used to generate the seed ensemble, the output $\Zout$ is later converted to normalised anomalies by setting
$\Xg_0=\Zout/\sqrt{\Ng-1}$ for $\Ng\ge 2$.

\begin{algorithm}[!htbp]
\caption{\label{alg:regular-simplex} Centred simplex ensemble with
prescribed total variance. $\qrQ$ returns the orthonormal factor of a reduced QR
decomposition. Indices start at $1$.}
\begin{algorithmic}[1]
  \setlength{\itemsep}{2.5pt}
\Require Integers $\Nout\ge 1,\Nx\ge 1$.
\Ensure $\Zout\in\Ex^{\Nout}$ for $\Nout\geq2$, or $\Zout=\bzero_{\Nx}$ for $\Nout=1$.
\Function{$[\Zout] =$ \textbf{regular\_simplex}}{$\Nout,\Nx$}
  \If{$\Nout=1$}
    \State \Return $\bzero_{\Nx}$
  \EndIf
  \State $k=\Nout-1$, \quad $m=\min(k,\Nx)$
  \State $\bC=[\bI_{\Nout}]_{:, 1:k}-\bone_{\Nout}\bone_k\T/\Nout$
  \State $\bOmega=\qrQ(\bC)$
  \State $\bX=\sqrt{\Nout/k}\,\bOmega\T$
  \If{$\Nx\geq k$}
    \State Draw $\bGamma\in\R^{\Nx\times k}$
      with $[\bGamma]_{i,j}\overset{\mathrm{iid}}{\sim}\gauss(0,1)$
    \State $\bG=\qrQ(\bGamma)$,
      \quad $\Zout=\bG\bX$
  \Else
    \State Draw $\bGamma\in\R^{k\times\Nx}$
      with $[\bGamma]_{i,j}\overset{\mathrm{iid}}{\sim}\gauss(0,1)$
    \State $\bG=\qrQ(\bGamma)$,
      \quad $\Zout=\bG\T\bX$
  \EndIf
  \State $\alpha=\sqrt{\Nx k^2/(m\Nout)}$
  \State $\Zout\gets\alpha\Zout$
  \State \Return $\Zout$
\EndFunction
\end{algorithmic}
\end{algorithm}

\subsection{Ensemble reduction and augmentation}

The resampling procedure in Algorithm~\ref{alg:redux-augment-rr} resizes a centred ensemble from $\Nin$ to $\Nout$
members by retaining its dominant covariance modes and randomly embedding them in the new ensemble space. Members occupy
the columns of the anomaly matrix $\Zin\in\Ex^{\Nin}$, with $\Zin\bone_{\Nin}=\bzero$. A thin singular value
decomposition identifies the numerical rank $r$ from the singular values $s_1\geq\cdots\geq s_m\geq0$. The value of $r$
is assessed through:
\begin{equation}
  \label{eq:rank}
  \rank(\bs) = \#\{i:s_i>\tau\}, \qquad \tau=\max(\Nin,\Nx)\,\epsilon_{\mathrm{mach}}\,s_1,
\end{equation}
with $\epsilon_{\mathrm{mach}}$ the machine numerical precision. Then, only the $q=\min(r,\Nout-1)$ leading modes are
retained.  An orthonormal basis of the subspace perpendicular to $\bone_{\Nout}$ is rotated using the orthogonal factor
of a Gaussian matrix, and its first $q$ columns embed the retained modes. The output is therefore centred. For ordinary
anomalies ($\nu=\mathrm{false}$), a scale factor ensures that $\Zout\Zout\T/(\Nout-1)$ is the rank-$q$ spectral
truncation of $\Zin\Zin\T/(\Nin-1)$, a best rank-at-most-$q$ approximation in both the spectral and Frobenius norms. For
normalised covariance factors ($\nu=\mathrm{true}$), no scaling is needed and the same statement holds for the unscaled
Gram matrices. Thus, the reduction retains the maximum variance among orthogonal projections of the input anomalies onto
subspaces of dimension at most $q$, whereas augmentation preserves the covariance without introducing new state-space
directions, up to the numerical-rank truncation and roundoff. Adding the original ensemble mean to the output ordinary
anomalies yields a resized ensemble with the same mean.

\begin{algorithm}[!htbp]
\caption{\label{alg:redux-augment-rr}
Ensemble reduction or augmentation with random rotation. $\qrQ$ returns the orthonormal factor of a reduced QR
decomposition. Indices start at $1$; a slice
$1:0$ is empty.}
\begin{algorithmic}[1]
\setlength{\itemsep}{2.5pt}
\Require Centred $\Zin\in\Ex^{\Nin}$;
$\Nx\geq1,\Nin\geq2,\Nout\geq2$; normalisation flag $\nu$.
\Ensure Centred $\Zout\in\Ex^{\Nout}$
in the input normalisation convention.
\Function{$[\Zout] =$ \textbf{SRA}}{$\Zin,\Nout,\nu$}
\State $m=\min(\Nx,\Nin)$
\State $(\bU,\bs,\bV\T)=\operatorname{svd}_{\mathrm{thin}}(\Zin)$
\State $r = \rank(\bs)$, as defined in Eq.~\eqref{eq:rank}.
\State $q=\min(r,\Nout-1)$
\State $\bB=\bU_{:,1:q}\diag(s_1,\ldots,s_q)$
\If{$q>0$}
    \State $\bC=\bI_{\Nout}-\bone_{\Nout}\bone_{\Nout}\T/\Nout$
    \State $\bG=\qrQ(\bC_{:,1:\Nout-1})$
    \State Draw $\bGamma\in\R^{(\Nout-1)\times(\Nout-1)}$
    with $[\bGamma]_{i,j}\overset{\mathrm{iid}}{\sim} \gauss(0,1)$
    \State $\bOmega=\qrQ(\bGamma)$,
    \quad $\bG\gets\bG\bOmega_{:,1:q}$
\Else
    \State $\bG=\bzero_{\Nout\times0}$
\EndIf
\State $c=\begin{cases}
    \nu=\mathrm{true}: & 1, \\
    \nu=\mathrm{false}: & \sqrt{(\Nout-1)/(\Nin-1)}, 
\end{cases}$
\State $\Zout=c\bB\bG\T$
\State \Return $\Zout$
\EndFunction
\end{algorithmic}
\end{algorithm}

\end{document}